\documentclass{article}

    \PassOptionsToPackage{numbers, compress}{natbib}

 \usepackage[main, final]{neurips_2026}

\usepackage[utf8]{inputenc} 
\usepackage[T1]{fontenc}    
\usepackage{hyperref}       
\usepackage{url}            
\usepackage{booktabs}       
\usepackage{amsfonts}       
\usepackage{nicefrac}       
\usepackage{microtype}      
\usepackage{xcolor}         
\usepackage{makecell}
\usepackage{amsmath}
\usepackage{graphicx}
\usepackage{caption}
\usepackage[most]{tcolorbox}
\usepackage{titlesec}
\usepackage{cleveref}
\usepackage{algorithm}
\usepackage{algpseudocode}
\usepackage{amsmath}
\usepackage{wrapfig}
\usepackage{enumitem}

\titlespacing*{\section}
{0pt}{0.8ex plus 0.2ex minus 0.1ex}{0.3ex}

\titlespacing*{\subsection}
{0pt}{0.5ex plus 0.1ex minus 0.1ex}{0.2ex}

\titlespacing*{\subsubsection}
{0pt}{0.4ex plus 0.1ex minus 0.1ex}{0.15ex}

\titlespacing*{\paragraph}
{0pt}{0.2ex}{0.3em}

\setlist[itemize,1]{
    leftmargin=1.5em,
    labelsep=0.45em,
    topsep=3pt,
    itemsep=2pt,
    parsep=0pt,
    partopsep=0pt
}

\title{FiRe: Fine-grained Multimodal Reasoning for Enhanced Image Generation}

\author{%
  Yongjin Kim$^{1}$ \quad Yoonjin Oh$^{1}$\thanks{Equal contribution.} \quad Yerin Kim$^{1}$\footnotemark[1] \quad Hyomin Kim$^{1}$ \\
  \textbf{Jeeyoung Yun}$^{1}$ \quad \textbf{Yu-Jung Heo}$^{2}$ \quad
  \textbf{Min-Jun Kim}$^{2}$ \quad \textbf{Sungwoong Kim}$^{1}$\thanks{Corresponding author.} \\[0.5em]
  $^{1}$Department of Artificial Intelligence, Korea University, Seoul, Republic of Korea \\
  $^{2}$KT Corporation, Seoul, Republic of Korea \\
  \texttt{\{rla020, dhdbsrlw, yerin-kim, khmiee, jee010910, swkim01\}@korea.ac.kr} \\
  \texttt{\{yj.heo, min\_jun.kim\}@kt.com}
}

\begin{document}

\maketitle

\begin{abstract}
    With the rapid progress of Multimodal Large Language Models (MLLMs), unified MLLMs that jointly perform image understanding and generation have advanced significantly. However, despite the inherent reasoning capabilities of unified MLLMs for self-reflection and self-refinement, their use in text-to-image generation remains largely underexplored. Meanwhile, existing multimodal reasoning–based image generation methods mostly rely on prompt augmentation or holistic image–text alignment judgments, without fine-grained reflection and refinement of detailed prompt attributes, leading to limited fine-grained control. To address this limitation, we propose \textbf{FiRe}, a \underline{\textbf{Fi}}ne-grained Multimodal \underline{\textbf{Re}}asoning method for enhanced image generation by MLLM. In specific, FiRe performs a fine-grained multi-step reasoning by first decomposing the prompt into key visual requirements and then self-judging their satisfaction in the generated image, followed by localized refinement according to self-generated precise feedback. In addition, to further strengthen the MLLM's multimodal reasoning ability, we introduce \textbf{FiRe-GRPO}, a reinforcement learning method tailored to FiRe. Since standard Group Relative Policy Optimization (GRPO) suffers from sparse, outcome-based rewards in multi-step reasoning, we formulate our reasoning process as a step-level decision-making problem, design step-specific rewards, and compute step-level advantages for granular credit assignment within GRPO. Extensive experiments demonstrate that FiRe consistently outperforms competitive text-to-image baselines, including existing reasoning-based methods, with particularly substantial gains on compositional text-to-image benchmarks. Our project page is available at \url{https://ku-agi.github.io/FiRe/}.
\end{abstract}

\section{Introduction}

\begin{figure}[t]
    \centering
    \includegraphics[width=1.0\linewidth]{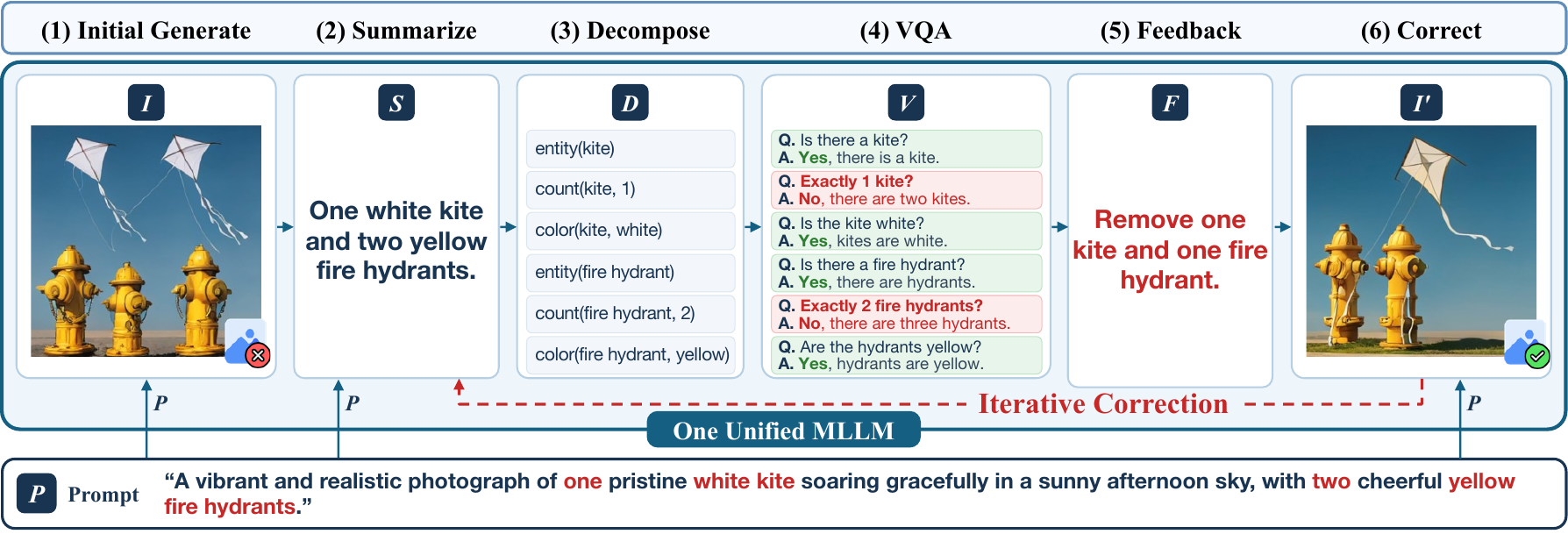}
    \caption{
        \textbf{Overview of FiRe.}
        FiRe iteratively refines the alignment between input prompts and output images through six steps.
        \textbf{Step~1:} generates an initial image from the input prompt.
        \textbf{Step~2:} summarizes the prompt into concrete visual details that can be checked in the image.
        \textbf{Step~3:} decomposes the summarized prompt into semantic tuples representing atomic visual elements.
        \textbf{Step~4:} verifies each tuple against the current image through tuple-level VQA.
        \textbf{Step~5:} generates fine-grained feedback for tuples judged as unsatisfied.
        \textbf{Step~6:} applies the feedback to perform localized image correction. This entire loop is repeated until the image--prompt alignment is fully satisfied.
        }
    \label{fig:Overview}
    \end{figure}

Recent advances in unified Multimodal Large Language Models (MLLMs)~\citep{chen2025janus,ma2025janusflow,ge2023making,ge2024seed,tong2025metamorph,xie2025show,ma2025unitok,wu2024vila,wang2024emu3} have enabled image understanding and image generation within a single model. However, reasoning has been much more actively explored for image understanding than for image generation.
In image understanding, reasoning techniques such as chain-of-thought prompting,
self-reflection, and self-refinement have been widely explored~\citep{zhang2025r1-vl,
xu2024r1-llava-cot, thawakar2025r1-llamaV-o1, yao2024mulberry,
zhang2024vlm-cot, huang2025visionr1}, enabling models to use additional
inference-time computation to reason, verify, and refine their answers. In contrast, reasoning for text-to-image (T2I) generation remains underexplored.
Although recent methods~\citep{pan2025janus, jiang2025t2i} introduce refinement
processes into T2I generation, they still struggle with complex compositional
prompts involving multiple objects, attributes, and spatial relations, as their
coarse holistic judgments cannot reliably verify fine-grained prompt
requirements.



To address this limitation, we propose \textbf{FiRe}, a \textbf{\underline{Fi}}ne-grained Multimodal \textbf{\underline{Re}}asoning method for Enhanced Image Generation with an MLLM. FiRe partitions the reasoning process into fine-grained steps to ensure that every compositional requirement is captured. As shown in~\Cref{fig:Overview}, FiRe consists of six reasoning steps: \textbf{(1) Initial T2I Generation}, \textbf{(2) Prompt Summarization}, \textbf{(3) Tuple Decomposition}, \textbf{(4) Tuple VQA}, \textbf{(5) Fine-grained Feedback Generation}, and \textbf{(6) Localized Image Correction}. 
FiRe first generates an initial image and summarizes the prompt by retaining concrete visual details while removing abstract or subjective descriptions. It then decomposes the summarized prompt into independently verifiable units and checks each unit through visual question answering (VQA). Based on the unsatisfied units, FiRe generates feedback and applies localized corrections only to mismatched regions. Through this fine-grained method, FiRe can identify unsatisfied compositional requirements that holistic judgment would otherwise miss.

To further improve fine-grained reasoning capability, we introduce FiRe-GRPO, a reinforcement learning method designed for FiRe. The standard Group Relative Policy Optimization (GRPO)~\citep{shao2024deepseekmath} treats the entire generation trajectory as a single action and assigns rewards only based on the quality of the final outcome. This outcome-based formulation often suffers from sparse feedback, as it fails to capture fine-grained signals from individual intermediate reasoning steps within the same trajectory. To address this limitation, FiRe-GRPO formulates our six-step reasoning process as a step-level decision-making problem. Since the six-step formulation enables a fine-grained self-evaluation process, FiRe-GRPO defines step-specific rewards and computes step-level advantages, rather than relying solely on a single outcome reward. This provides targeted learning signals for intermediate reasoning decisions, including prompt summarization, tuple decomposition, tuple-level VQA, feedback generation, and localized image correction.

Experiments on representative compositional T2I benchmarks, including GenEval~\citep{ghosh2023geneval}, GenEval++~\citep{ye2025echo}, and DPG-Bench~\citep{hu2024ella}, demonstrate that FiRe consistently outperforms existing baselines. Our main contributions are summarized as follows:

\begin{itemize}
    \item We propose FiRe, a fine-grained multimodal reasoning method for text-to-image generation that identifies prompt--image misalignments at a fine-grained level and corrects only the mismatched regions through localized refinement.

    \item We introduce FiRe-GRPO, a step-level reinforcement learning method for FiRe that assigns rewards according to the role of each reasoning step and computes step-level advantages, enabling fine-grained credit assignment to intermediate judgments, feedback, and corrections.

    \item We conduct extensive experiments and ablation studies on compositional T2I benchmarks, showing that FiRe consistently outperforms strong image generation and reasoning-based baselines.
\end{itemize}

\section{Method}

We first define the six fine-grained reasoning steps of FiRe. We then describe supervised fine-tuning for learning this reasoning process, followed by FiRe-GRPO, a step-level reinforcement learning method that further strengthens step-wise reasoning ability. All six steps are performed by a single unified MLLM, without any external generator, editor, or evaluator at inference. External models are used only during training, for SFT data construction (\Cref{sec2.2:SFT}) and reward computation in FiRe-GRPO (\Cref{sec2.3:GRPO}). ~\Cref{fig:Overview} provides an overview of FiRe, and we detail each reasoning step below.

\subsection{Fine-grained Reasoning Steps in FiRe}
\label{reasoning_step_subsection}


\paragraph{Step 1: Initial Text-to-Image Generation.}
FiRe starts by generating an initial image $I$ from the input prompt $p$. The initial image is expected to be aligned with the prompt, but may still miss or incorrectly render fine-grained visual details. FiRe therefore uses $I$ as the starting point for the subsequent steps.

\paragraph{Step 2: Prompt Summarization.}
After generating the initial image $I$, FiRe summarizes the input prompt $p$ into $S$ by retaining concrete visual details that can be explicitly checked in the image, such as objects, attributes, counts, spatial relations, and text, while removing subjective or non-verifiable intangible descriptions, such as \textit{beautiful}, \textit{vibrant}, or \textit{aesthetic}.

\paragraph{Step 3: Tuple Decomposition.}
FiRe then decomposes $S$ into semantic tuples $D=\{\tau_i\}_{i=1}^{Q}$. Each tuple $\tau_i$ captures an atomic visual element, such as an entity, attribute, relation, count, spatial constraint, or text element. For example, if the summarized prompt $S$ contains ``a pink oven'', FiRe decomposes it into \texttt{whole-entity(oven)} and \texttt{attribute-color(oven, pink)}.

\paragraph{Step 4: Tuple VQA.}
Each semantic tuple in $D$ is checked against the current image $I$ through tuple-level VQA, producing tuple VQA results $V = \{v_i\}_{i=1}^{Q}$, where $v_i = (r_i, y_i)$.
Each result $v_i$ contains an image-grounded rationale $r_i$ and a binary judgment $y_i\in\{\texttt{yes},\texttt{no}\}$ indicating whether the visual requirement represented by $\tau_i$ is satisfied. This step converts each tuple into an explicit verification result. For example, suppose the prompt requires a pink oven, while the generated image contains a white oven. For the tuple $\tau_i=\texttt{attribute-color(oven, pink)}$, FiRe asks the verification question ``Is the oven pink?'' and may produce $(r_i,y_i)=(\text{``The oven is white rather than pink.''},\texttt{no})$.


\paragraph{Step 5: Fine-grained Feedback Generation.}
Using the tuple-level verification results
$V=\{(r_i,y_i)\}_{i=1}^{Q}$, FiRe generates feedback $F$ based on whether the
tuples are satisfied. 
If all tuples are satisfied, FiRe terminates the correction process with $F=\texttt{no-edit}$;
otherwise, it generates actionable feedback $F=\tilde{F}$ for correcting the unsatisfied tuples $\{\tau_i \mid y_i=\texttt{no}\}$. When
multiple tuples are unsatisfied, $\tilde{F}$ combines the required corrections
into a single feedback instruction so that all identified mismatches are
addressed together. For example, if the prompt requires \texttt{attribute-color(oven, pink)} and
\texttt{count(plate, 3)}, but the generated image contains a white oven and only
two plates, FiRe judges both requirements as \texttt{no} and may produce
$\tilde{F}=\text{``Change the oven color from white to pink and add one more
plate to make three plates.''}$

\paragraph{Step 6: Localized Image Correction.}
When $F\neq\texttt{no-edit}$, FiRe applies the feedback $F$ to the current image
$I$ and produces a corrected image $I'$. This localized correction is performed
using an editing prompt that includes $F$ together with an explicit preservation
instruction, thereby targeting the unsatisfied tuples identified in the previous
steps while preserving already satisfied content. FiRe then sets
$I\leftarrow I'$ and uses the corrected image as the input to Step~2 in the next
correction attempt.

Detailed input prompts and FiRe's inference algorithm are provided in Appendices~\ref{app:fire_prompts} and~\ref{app:fire_inference}.

\subsection{Supervised Fine-tuning for Multimodal Reasoning}
\label{sec2.2:SFT}

Starting from a pretrained unified MLLM~\citep{chen2025janus} $\pi_\theta$, we first use supervised fine-tuning (SFT) to equip the model with FiRe's fine-grained multimodal reasoning ability. We construct an SFT dataset $\mathcal{D}_{\mathrm{SFT}}$ consisting of supervised FiRe trajectories from both prompt-aligned and prompt-misaligned initial images. If the initial image $I$ is already aligned with the prompt, the trajectory contains the intermediate reasoning outputs and ends with $F=\texttt{no-edit}$; otherwise, it further includes fine-grained feedback $F$ and a corrected image $I'$:
\begin{equation}
(I,S,D,V,F)\in\mathcal{D}_{\mathrm{SFT}}
\quad
\text{if } F=\texttt{no-edit},
\qquad
(I,S,D,V,F,I')\in\mathcal{D}_{\mathrm{SFT}}
\quad
\text{otherwise}.
\end{equation}
Here, $I$ is the initial image, $S$ is the summarized prompt, $D$ is the tuple decomposition, $V$ is the tuple-level VQA result, $F$ is the feedback, and $I'$ is the corrected image.

We train the model using token-level cross-entropy over the supervised FiRe
output sequences. For each supervised output sequence $Y=(y_1,\ldots,y_T)$, let $X$ denote the corresponding input context. For
example, $X=p$ for initial image generation in Step~1, $X=(p,I)$ for the
fine-grained reasoning trace in Steps~2--5, and $X=(p,I,F)$ for localized image
correction in Step~6. The SFT loss is
\begin{equation}
\mathcal{L}_{\mathrm{SFT}}(\theta)
=
-\frac{1}{T}
\sum_{\ell=1}^{T}
\log \pi_\theta(y_\ell \mid X,y_{<\ell}),
\end{equation}
where $T$ denotes the target sequence length. The data preparation procedure used to construct $\mathcal{D}_{\mathrm{SFT}}$ is provided in Appendix~\ref{app:data_preparation_details}.

\begin{figure}[t]
    \centering
    \includegraphics[width=1.0\linewidth]{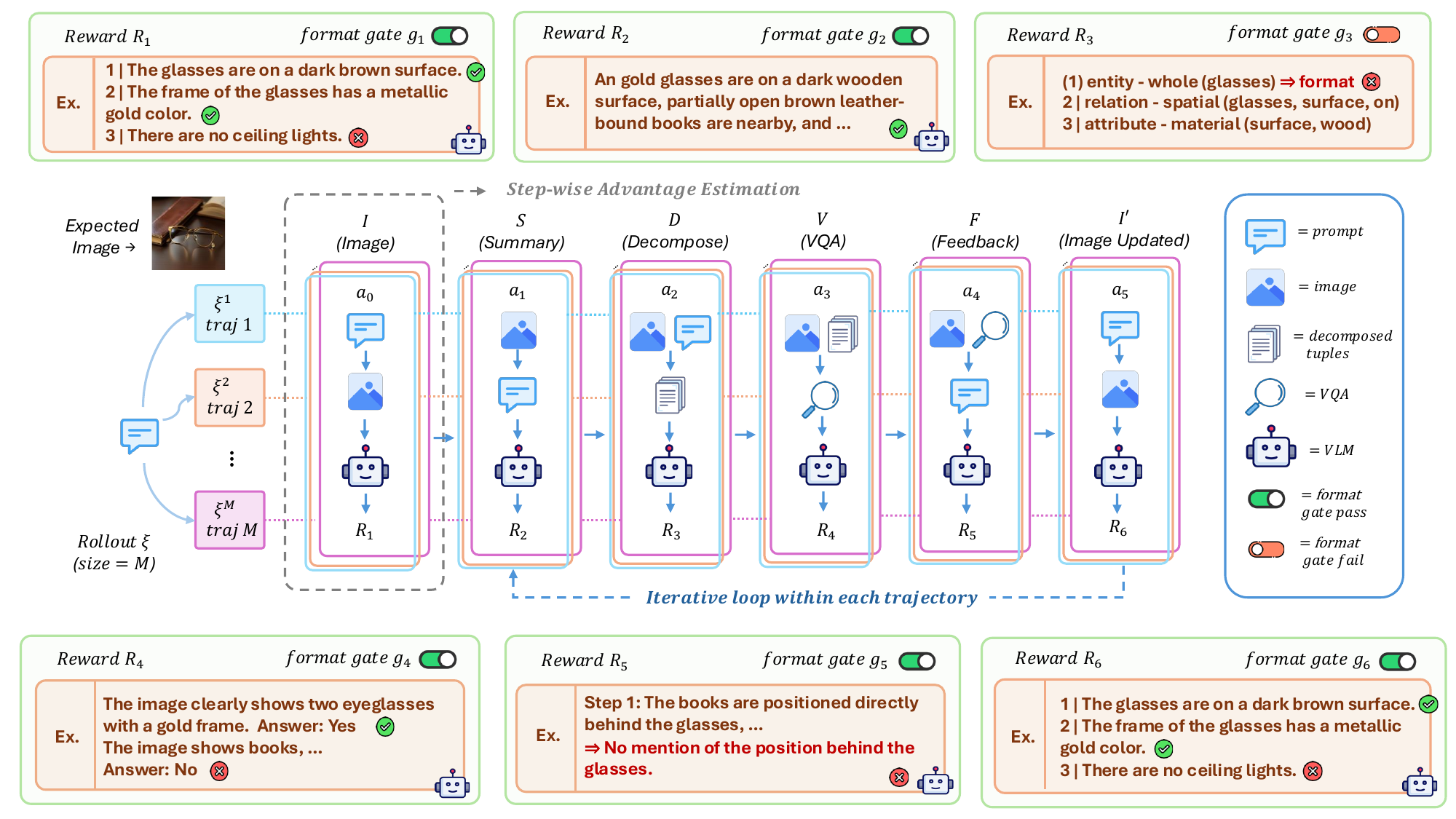}
    \caption{\textbf{Overview of FiRe-GRPO}. FiRe-GRPO performs step-wise advantage estimation at the level of individual FiRe reasoning steps. At each step, the policy samples a group of outputs, and the reward model assigns a step-specific reward to each output. For Steps 2--5, a format gate sets the reward to zero when the output violates the required structured format. The advantages are then computed within the corresponding step-level reward group.}
    \label{fig:FiRe_Reward}
\end{figure}

\subsection{FiRe-GRPO: Step-level Reinforcement Learning}
\label{sec2.3:GRPO}

Although SFT equips the model with FiRe's fine-grained reasoning process, it learns by imitating supervised trajectories rather than optimizing rewards over trajectories sampled from the model's own policy. To extend FiRe beyond SFT, we introduce FiRe-GRPO represented by ~\Cref{fig:FiRe_Reward}, a reinforcement learning method that optimizes FiRe using step-specific rewards and step-level advantages.

\paragraph{Step-level Sequential Decision-making Formulation.}
To formulate FiRe as a sequential decision-making problem, we define each action at the level of a step output. Namely, an
output action in FiRe by $\pi_{\theta}$ can be the initial image $I$, summarized prompt $S$, tuple decomposition
$D$, tuple-level VQA results $V$, feedback $F$, or corrected image $I'$.

At each action (decision) position $h$ corresponding to each step, the state $s_h$ is the multimodal context used
to generate the current step output. It contains the input prompt, the current image, and the FiRe outputs generated
in previous steps, from which the policy samples a step-level action
$a_h \sim \pi_\theta(\cdot \mid s_h)$.

Since FiRe may repeat Steps~2--6 over multiple correction attempts, a rollout
contains a sequence of such step-level actions. Given a prompt $p$, the policy $\pi_\theta$
samples a group of $M$ FiRe rollouts $\{\xi^{(i)}\}_{i=1}^{M}$. We index the actions in each rollout by their generation order,
$h=0,\ldots,H_i-1$, and write
$\xi^{(i)}=\{(s_h^{(i)},a_h^{(i)},r_h^{(i)})\}_{h=0}^{H_i-1}$.
where, for each action $a_h^{(i)}$ that is associated with a step label $k\in\{1,\ldots,6\}$ (as defined in~\Cref{reasoning_step_subsection}), $r_h^{(i)}=R_k(s_h^{(i)},a_h^{(i)})$ is the step-level reward computed by the corresponding reward function $R_k$.

\subsubsection{Step-specific Reward Design}

Each FiRe step has a specific role in the reasoning process, and the model
should learn to perform each step according to its intended function. Existing
GRPO-based methods often optimize generated outputs using a scalar reward
computed from the final outcome~\citep{jiang2025t2i,shao2024deepseekmath}.
However, a final-image-only reward is insufficient for FiRe: it does not provide
direct training signals for intermediate reasoning outputs, such as the prompt
summary, tuple decomposition, tuple-level verification, and feedback. We
therefore design step-level rewards using three reward components: 
\textbf{fine-grained image alignment reward}, 
\textbf{textual reasoning evaluation reward}, and 
\textbf{instruction-following reward}. We use an off-the-shelf VLM-based evaluator
$\mathcal{E}_{\mathrm{vlm}}$ for reward computation.


\paragraph{Fine-grained image alignment reward. \quad}

For Steps~1 and~6, we evaluate image--prompt alignment at a fine-grained level.
Instead of asking a single binary question such as ``Does the image match the
\{prompt\}?'', the evaluator follows the same procedure as FiRe's Steps~2--4: it
summarizes the prompt, decomposes it into atomic visual elements, and converts
them into verification questions. Specifically, the evaluator first summarizes the prompt $p$ into an
evaluator-side summary $S^\star$, and then decomposes $S^\star$ into
$D^\star=\{\tau_i^\star\}_{i=1}^{Q}$. Each tuple $\tau_i^\star$ is converted
into a verification question $q_i^\star$. For example, $\tau_i^\star=\texttt{attribute-color(oven, pink)}$ can be converted into $q_i^\star=\text{``Is the oven pink?''}$. We use $\star$ to indicate quantities produced by the evaluator. The image alignment reward is defined as the fraction of questions answered as \texttt{yes} among the total $Q$ questions, i.e., $z_{\mathrm{img}}(I,p)=\frac{1}{Q}\sum_{i=1}^{Q}\mathbf{1}[\mathcal{E}_{\mathrm{vlm}}(I,q_i^\star)=\texttt{yes}]$.
This score $z$ provides a fractional score that captures which specific visual requirements are satisfied or violated, rather than only measuring a single binary image--prompt alignment.

\paragraph{Textual reasoning evaluation reward. \quad}
For Steps~2--5, the evaluator scores each output according to the
function of its corresponding step. Step~2 evaluates whether the summarized
prompt preserves checkable visual details while removing subjective or
non-verifiable descriptions. Step~3 evaluates whether the tuple decomposition
faithfully captures atomic visual elements without omission or hallucination.
Step~4 evaluates whether the tuple-level VQA results provide image-grounded
rationales and accurate \texttt{yes}/\texttt{no} judgments. Step~5 evaluates
whether the feedback targets tuples judged as \texttt{no} and provides
actionable correction guidance without unnecessary changes to already satisfied
content. For these steps, we additionally apply a format gate,
which assigns zero reward when the generated output does not follow the
required step-specific structure, such as parseable tuples for Step~3,
rationale with \texttt{yes}/\texttt{no} judgments for Step~4, and
\texttt{no-edit} or actionable feedback for Step~5. This is necessary because
these structured outputs are used as inputs to subsequent FiRe steps.

\paragraph{Instruction-following reward. \quad}
For Step~6, we use the fine-grained image alignment reward for the corrected
image $I'$ and additionally incorporate an instruction following reward. The
instruction-following reward evaluates whether the edit from the current image
$I$ to the corrected image $I'$ faithfully follows the generated feedback $F$,
while avoiding unnecessary modifications to visual content that already satisfied the prompt requirements.

\paragraph{Reward assignment. \quad}
We assign a step-specific reward to each generated FiRe step output $a^{(i)}_h$ based on its step index $k$. Step~1 uses a fine-grained image alignment reward, Steps~2--5 use
role-specific rewards with format gating, and Step~6 uses both a fine-grained image
alignment reward and an instruction following reward. For example, when $k=1$,
$a^{(i)}_h$ is the initial image $I^{(i)}$, and the assigned reward is
$
r^{(i)}_h
=
R_{k}(s^{(i)}_h,a^{(i)}_h)
=
R_1(s^{(i)}_h,a^{(i)}_h)
=
z_{\mathrm{img}}(I^{(i)},p).$

Detailed reward computation, required formats, and evaluator prompts are
provided in Appendix~\ref{app:reward}.

\subsubsection{Stepwise Advantage Estimation}

Aggregating all step-specific rewards into a single outcome reward
and assigning one uniform advantage to the entire rollout would 
obscure which FiRe steps were more or less effective.
FiRe-GRPO therefore computes advantages at the level of FiRe reasoning components, preserving step-aware credit assignment.

For each generation position $h$, we compute the advantage of action
$a_h^{(i)}$ using its step-specific reward $r_h^{(i)}$. Given a rollout group
sampled for the same prompt, the reward is normalized across the $M$ rollouts at
the same position $h$:
\begin{equation}
A_h^{(i)}
=
\frac{
r_h^{(i)}
-
\operatorname{mean}\left(\{r_h^{(j)}\}_{j=1}^{M}\right)
}{
\operatorname{std}\left(\{r_h^{(j)}\}_{j=1}^{M}\right)
}.
\end{equation}

\subsubsection{FiRe-GRPO Optimization Objective}

Given a prompt $p \sim \mathcal{D}_{\mathrm{GRPO}}$, where $\mathcal{D}_{\mathrm{GRPO}}$ denotes the training prompt distribution, we sample a group of $M$ rollouts
$\{\xi^{(i)}\}_{i=1}^{M}$ from the old policy $\pi_{\theta_{\mathrm{old}}}(\cdot \mid p)$. For each rollout $\xi^{(i)}$, we compute the step-specific rewards and the corresponding advantages as described above. We then incorporate our step-level advantages into the token-level GRPO objective~\citep{shao2024deepseekmath}. In particular, for each rollout $\xi^{(i)}$, let $T_i$ denote the total number of generated
tokens across all FiRe step outputs, and let $z_t^{(i)}$ and $c_t^{(i)}$ be the $t$-th generated token and the autoregressive
context used to generate $z_t^{(i)}$, respectively. The token-level importance ratio is $\rho_t^{(i)}(\theta)
=
\frac{
\pi_\theta(z_t^{(i)}\mid c_t^{(i)})
}{
\pi_{\theta_{\mathrm{old}}}(z_t^{(i)}\mid c_t^{(i)})
}$.
Then, we define the token-level clipped objective as
\begin{equation}
j_t^{(i)}(\theta)
=
\min
\left(
\rho_t^{(i)}(\theta)\widehat{A}_t^{(i)},
\operatorname{clip}
\left(
\rho_t^{(i)}(\theta),
1-\epsilon,
1+\epsilon
\right)
\widehat{A}_t^{(i)}
\right),
\end{equation}
where $\epsilon$ is the hyperparameter for clipping. Here, it is noted that if a token $z_t^{(i)}$ is generated within action $a_h^{(i)}$, we set
$\widehat{A}_t^{(i)}=A_h^{(i)}$.
Finally, the FiRe-GRPO objective is
\begin{align}
\mathcal{J}_{\mathrm{FiRe\text{-}GRPO}}(\theta)
=
\mathbb{E}_{p,\{\xi^{(i)}\}}
\!\left[
\frac{1}{M}
\sum_{i=1}^{M}
\frac{1}{T_i}
\sum_{t=1}^{T_i}
\left(
j_t^{(i)}(\theta)
-
\beta
D_{\mathrm{KL}}
\bigl(
\pi_\theta(\cdot|c_t^{(i)})
\,\|\,
\pi_{\mathrm{ref}}(\cdot|c_t^{(i)})
\bigr)
\right)
\right],
\end{align}
where $\beta$ and $\pi_{\mathrm{ref}}$ indicate the coefficient and the reference (SFT) model for the KL penalty, respectively. Our MLLM for FiRe, $\pi_\theta$, is optimized by maximizing
$\mathcal{J}_{\mathrm{FiRe\text{-}GRPO}}(\theta)$.

\begin{table*}[t!]
\centering
\caption{Comparison on GenEval \citep{ghosh2023geneval}, GenEval++ \citep{ye2025echo}, and DPGBench \citep{hu2024ella}. ↑ indicates higher is better. \textbf{Bold} and \underline{underline} denote the best and second-best results among Multimodal Reasoning LLMs, respectively. Superscript * indicates results reproduced by us. (1st) denotes the initial T2I generation, while (2nd) and (3rd) represent the first and second rounds of image correction, respectively. Subcategory scores of DPGBench are provided in Appendix~\ref{app:additional_quantitative}.}
{\fontsize{7}{9}\selectfont
\setlength{\tabcolsep}{0.3mm}
\begin{tabular}{lccccccc c cccccccc c c}
\toprule
& \multicolumn{7}{c}{\textbf{GenEval ↑}} &
& \multicolumn{8}{c}{\textbf{GenEval++ ↑}} &
& \multicolumn{1}{c}{\textbf{DPGBench ↑}} \\
\cline{2-8}\cline{10-17}\cline{19-19}
\textbf{Method} &
Overall & Single & Two & Count & Color & Pos & Color Attr &
\multicolumn{1}{c}{} &
Overall & Color & Count & \makecell{Color/\\Count} & \makecell{Color/\\Pos} & \makecell{Pos/\\Count} & \makecell{Pos/\\Size} & \makecell{Multi-\\Count} &
\multicolumn{1}{c}{} &
Overall \\
\midrule
\multicolumn{19}{c}{\textbf{Diffusion Models}} \\ \midrule
SD3.5 Large \citep{esser2024scaling}     & 0.74 & 0.99 & 0.94 & 0.72 & 0.89 & 0.33 & 0.60
& & - & - & - & - & - & - & - & - & & 84.08 \\
Flux.1 Dev \citep{flux2024}      & 0.66 & 0.98 & 0.79 & 0.73 & 0.77 & 0.22 & 0.45 
& & 0.314 & 0.350 & 0.625 & 0.150 & 0.275 & 0.200 & 0.375 & 0.225 & & 83.79 \\
\midrule
\multicolumn{19}{c}{\textbf{Unified Multimodal Large Language Models}} \\
\midrule
Emu3 \citep{wang2024emu3}    & 0.52 & 0.98 & 0.69 & 0.33 & 0.78 & 0.15 & 0.16 
& & - & - & - & - & - & - & - & - & & 81.60 \\
Show-o2 \citep{xie2025show}   & 0.76 & 1.00 & 0.87 & 0.58 & 0.92 & 0.52 & 0.62 
& & - & - & - & - & - & - & - & - & & 86.14 \\
Unitok \citep{ma2025unitok}     & 0.59 & 0.99 & 0.75 & 0.39 & 0.80 & 0.21 & 0.41 
& & - & - & - & - & - & - & - & - & & 81.87 \\
Blip-3o \citep{chen2025blip3}   & 0.84 & - & - & - & - & - & - 
& & 0.307 & 0.250 & 0.250 & 0.125 & 0.600 & 0.125 & 0.575 & 0.225 & & 81.60 \\
BAGEL \citep{deng2025emerging}   & 0.82 & 0.99 & 0.94 & 0.81 & 0.88 & 0.64 & 0.63 
& & 0.371 & 0.325 & 0.600 & 0.250 & 0.325 & 0.250 & 0.475 & 0.375 & & 85.07 \\
Janus-Pro-7B    & 0.80 & 0.99 & 0.89 & 0.59 & 0.90 & 0.79 & 0.66
& & 0.246 & 0.450 & 0.300 & 0.125 & 0.300 & 0.075 & 0.350 & 0.125 & & 83.78 \\
\midrule
\multicolumn{19}{c}{\textbf{Multimodal Reasoning Models}} \\
\midrule
T2I-R1$^{*}$ \citep{jiang2025t2i}
& 0.79 & \textbf{1.00} & 0.91 & 0.52 & 0.90 & 0.77 & 0.65
& & 0.311 & 0.675 & 0.325 & 0.200 & 0.350 & 0.075 & 0.250 & 0.300
& & 85.06 \\

Janus-R1$^{*}$ (1st) \citep{pan2025janus}
& 0.80 & \textbf{1.00} & 0.92 & 0.48 & 0.91 & 0.80 & 0.71
& & 0.290 & 0.481 & 0.306 & 0.244 & 0.369 & 0.094 & 0.300 & 0.238
& & 83.58 \\

Janus-R1$^{*}$ (2nd)
& 0.82 & \textbf{1.00} & \underline{0.93} & 0.51 & 0.92 & 0.87 & 0.70
& & 0.294 & 0.481 & 0.369 & 0.200 & 0.363 & 0.113 & 0.313 & 0.219
& & 84.03 \\

Janus-R1$^{*}$ (3rd)
& 0.83 & \textbf{1.00} & \textbf{0.94} & 0.51 & \underline{0.93} & \underline{0.88} & 0.72
& & 0.285 & 0.481 & 0.281 & 0.200 & 0.344 & 0.150 & 0.306 & 0.231
& & 84.13 \\

\midrule

\textbf{FiRe} (1st)
& 0.84 & \underline{0.99} & \textbf{0.94} & 0.59 & \underline{0.93} & 0.85 & 0.77
& & 0.379 & 0.625 & 0.325 & 0.275 & 0.550 & 0.175 & 0.369 & 0.331
& & 84.88 \\

\textbf{FiRe} (2nd)
& 0.86 & \underline{0.99} & \textbf{0.94} & 0.70 & \underline{0.93} & 0.85 & 0.77
& & 0.464 & 0.750 & 0.444 & 0.381 & 0.575 & 0.263 & 0.375 & 0.463
& & 85.26 \\

\textbf{FiRe} (3rd)
& 0.87 & \underline{0.99} & \textbf{0.94} & 0.72 & \underline{0.93} & 0.85 & 0.77
& & 0.491 & \textbf{0.775} & 0.488 & 0.450 & 0.569 & 0.281 & 0.356 & 0.488
& & 85.28 \\

\midrule

\textbf{FiRe-2500 Steps} (1st)
& \underline{0.89} & \textbf{1.00} & \textbf{0.94} & \underline{0.77} & \textbf{0.94} & \textbf{0.90} & \underline{0.79}
& & 0.591 & 0.738 & 0.413 & 0.569 & \underline{0.788} & 0.475 & \textbf{0.756} & 0.400
& & 85.40 \\

\textbf{FiRe-2500 Steps} (2nd)
& \textbf{0.90} & \textbf{1.00} & \textbf{0.94} & \textbf{0.81} & \textbf{0.94} & \textbf{0.90} & \textbf{0.81}
& & \underline{0.684} & 0.731 & \underline{0.550} & \textbf{0.700} & \underline{0.788} & \textbf{0.688} & \underline{0.750} & \underline{0.581}
& & \underline{85.97} \\

\textbf{FiRe-2500 Steps} (3rd)
& \textbf{0.90} & \textbf{1.00} & \textbf{0.94} & \textbf{0.81} & \textbf{0.94} & \textbf{0.90} & \textbf{0.81}
& & \textbf{0.686} & \underline{0.756} & \textbf{0.581} & \underline{0.681} & \textbf{0.798} & \underline{0.638} & \textbf{0.756} & \textbf{0.594}
& & \textbf{86.01} \\

\bottomrule
\end{tabular}
}
\label{tab:geneval_genevalpp}
\end{table*}
\section{Experiments}
\label{}

\begin{figure*}[t!]
    \centering
    \includegraphics[width=\textwidth]{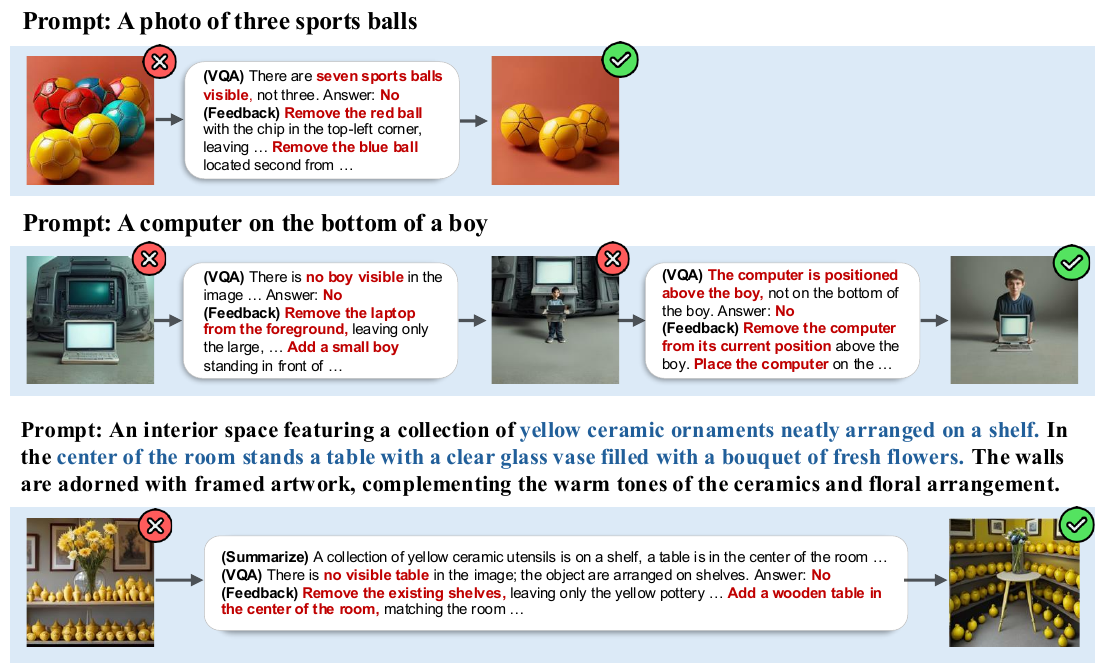}
    \caption{Qualitative Results of FiRe. From left to right, the first image denotes the initial T2I generation, the second image displays the result following the first round of image correction, and the rightmost image represents the output after the second round of image correction.
}
    \label{fig:qualitative_comp} 
\end{figure*}

\subsection{Experimental Setup}
\label{sec:experimental_setup}

\paragraph{Backbone and baselines. \quad}
We use Janus-Pro-7B~\citep{chen2025janus} as the backbone for FiRe and compare against two reasoning-based baselines built on the same backbone: Janus-Pro-R1~\citep{pan2025janus}, which performs holistic image assessment followed by full-image regeneration, and T2I-R1~\citep{jiang2025t2i}, which augments prompts via text-centric chain-of-thought reasoning prior to generation. We use Qwen3.5-35B-A3B~\citep{qwen35blog} as the reward evaluator for FiRe-GRPO. 

\paragraph{Training data. \quad}

For supervised fine-tuning, we use 200K image-text pairs: 140K from FocusDiff~\citep{pan2025focusdiff} and 60K composition-specific samples synthesized with Qwen-Image and Qwen-Image-Edit~\citep{wu2025qwen}. The prompts for the synthesized samples are adapted from Flow-GRPO~\citep{liu2025flow} and ReasonGen-R1~\citep{zhang2025reasongen}. We additionally incorporate complex and long prompts from Echo-4o-Image~\citep{ye2025echo} and LongAlign~\citep{liu2024improving}.

\paragraph{Benchmarks. \quad}
We evaluate FiRe on GenEval~\citep{ghosh2023geneval}, GenEval++~\citep{ye2025echo}, and DPGBench~\citep{hu2024ella}. Detailed descriptions of the benchmarks and evaluation protocols are provided in Appendix~\ref{app:benchmark_details}. To ensure a fair comparison, we reproduce the results of all multimodal reasoning baseline models in the same environment under identical evaluation settings. Additional details are provided in Appendix~\ref{app:implementation_details}.

\subsection{Quantitative Results}

Table~\ref{tab:geneval_genevalpp} compares FiRe with state-of-the-art diffusion
models, unified MLLMs, and multimodal reasoning models.
Among the baselines, only Janus-Pro-R1 and FiRe support iterative refinement.
Therefore, we report their initial generation (1st) and two subsequent refinement stages (2nd and 3rd).
At each refinement stage, the model edits the image only when its self-judgment indicates a prompt--image mismatch; otherwise, the previous image is retained.
Notably, due to the substantial computational cost of separately training each
FiRe-GRPO variant, all FiRe-GRPO results in the standard evaluation and
ablation studies use the 300-step checkpoint described in
Appendix~\ref{app:training_details}.
We additionally report FiRe-2500 Steps, obtained by continuing the same
FiRe-GRPO training to 2,500 steps. Overall, FiRe achieves the best performance across nearly all benchmarks, with further improvements from extended training.

\subsection{Qualitative Results}
Figure~\ref{fig:qualitative_comp} presents representative FiRe results on
different prompts. Across diverse compositional prompts, FiRe
identifies fine-grained misalignments and selectively edits erroneous regions
while preserving correctly aligned content. These examples show that iterative,
fine-grained reasoning improves prompt--image alignment. Additional qualitative
results are provided in Appendix~\ref{app:additional_qualitative}.

\subsection{Ablation Study}
\label{ablation}
\begin{table*}[t!]
\centering

\begin{minipage}[t]{0.58\linewidth}
\centering
{\fontsize{8}{9.5}\selectfont
\setlength{\tabcolsep}{0.6mm}

\captionof{table}{
Ablation on reasoning step granularity.
Comparison of fine-grained multi-step (w/ FG) and coarse single-step
(w/ Coarse) reasoning across three refinement iterations on GenEval++.
}
\label{tab:ablation1}

\resizebox{\linewidth}{!}{%
\begin{tabular}{lcccccccc}
\toprule
& \multicolumn{8}{c}{\textbf{GenEval++ $\uparrow$}} \\
\cmidrule{2-9}
\textbf{Method}
& Overall
& Color
& Count
& \makecell{Color/\\Count}
& \makecell{Color/\\Pos}
& \makecell{Pos/\\Count}
& \makecell{Pos/\\Size}
& \makecell{Multi-\\Count} \\
\midrule

\textbf{w/ FG} (1st)
& 0.35 & 0.58 & 0.28 & 0.24 & 0.49 & 0.11 & \textbf{0.39} & 0.36 \\

\textbf{w/ FG} (2nd)
& \underline{0.41}
& \underline{0.66}
& 0.39
& \underline{0.30}
& \underline{0.54}
& 0.16
& \textbf{0.39}
& \underline{0.42} \\

\textbf{w/ FG} (3rd)
& \textbf{0.44}
& \textbf{0.68}
& \underline{0.44}
& \textbf{0.34}
& \textbf{0.56}
& 0.20
& \textbf{0.39}
& \textbf{0.46} \\

\midrule

w/ Coarse (1st)
& 0.30 & 0.53 & 0.29 & 0.24 & 0.40 & 0.13 & 0.24 & 0.29 \\

w/ Coarse (2nd)
& 0.40
& 0.64
& \underline{0.44}
& 0.27
& \underline{0.54}
& \underline{0.22}
& \underline{0.27}
& 0.39 \\

w/ Coarse (3rd)
& \underline{0.41}
& 0.64
& \textbf{0.49}
& 0.29
& \underline{0.54}
& \textbf{0.23}
& 0.26
& 0.39 \\

\bottomrule
\end{tabular}
}%
}
\end{minipage}
\hfill
\begin{minipage}[t]{0.40\linewidth}
\centering
{\fontsize{9}{10.5}\selectfont
\setlength{\tabcolsep}{1.2mm}

\caption{Ablation of the Tuple Decomposition and VQA steps in FiRe-SFT on GenEval++, showing their contributions across three iterative refinement stages.}
\label{tab:fire_sft_ablation}

\resizebox{\linewidth}{!}{%
\begin{tabular}{lccc}
\toprule
\textbf{Method} & \textbf{1st} & \textbf{2nd} & \textbf{3rd} \\
\midrule
FiRe-SFT
& 0.348
& \textbf{0.409}
& \textbf{0.440} \\

w/o Tuple
& \textbf{0.361}
& 0.403
& 0.433 \\

w/o VQA
& 0.353
& 0.401
& 0.428 \\

w/o Tuple+VQA
& 0.337
& 0.387
& 0.407 \\
\bottomrule
\end{tabular}
}%
}
\end{minipage}

\end{table*}
\begin{table*}[t!]
\centering

\begin{minipage}[t]{0.49\linewidth}
\centering
{\fontsize{7.5}{9.5}\selectfont
\setlength{\tabcolsep}{0.4mm}

\captionof{table}{
Ablation on advantage estimation strategies in FiRe-GRPO on GenEval++.
}
\label{tab:ablation2}

\resizebox{\linewidth}{!}{%
\begin{tabular}{lcccccccc}
\toprule
& \multicolumn{8}{c}{\textbf{GenEval++ $\uparrow$}} \\
\cmidrule{2-9}
\textbf{Method}
& Overall
& Color
& Count
& \makecell{Color/\\Count}
& \makecell{Color/\\Pos}
& \makecell{Pos/\\Count}
& \makecell{Pos/\\Size}
& \makecell{Multi-\\Count} \\
\midrule

\textbf{w/ Stepwise} (1st)
& 0.38 & 0.63 & 0.33 & 0.28 & 0.55 & 0.18 & 0.37 & 0.33 \\

\textbf{w/ Stepwise} (2nd)
& \underline{0.46}
& \underline{0.75}
& 0.44
& 0.38
& \textbf{0.58}
& \underline{0.26}
& 0.38
& \underline{0.46} \\

\textbf{w/ Stepwise} (3rd)
& \textbf{0.49}
& \textbf{0.78}
& \underline{0.49}
& \textbf{0.45}
& \underline{0.57}
& \textbf{0.28}
& 0.36
& \textbf{0.49} \\

\midrule

w/ Aggregated (1st)
& 0.37 & 0.55 & 0.33 & 0.28 & 0.53 & 0.18 & \textbf{0.44} & 0.26 \\

w/ Aggregated (2nd)
& 0.39 & 0.61 & 0.38 & 0.29 & 0.51 & 0.19 & \textbf{0.44} & 0.31 \\

w/ Aggregated (3rd)
& 0.44 & 0.64 & 0.48 & 0.36 & \textbf{0.58} & 0.23 & \textbf{0.44} & 0.40 \\

\midrule

w/ Terminal (1st)
& 0.35 & 0.60 & 0.28 & 0.24 & 0.51 & 0.14 & 0.37 & 0.32 \\

w/ Terminal (2nd)
& 0.42 & 0.63 & 0.43 & 0.33 & 0.55 & \underline{0.26} & 0.38 & 0.37 \\

w/ Terminal (3rd)
& 0.44 & 0.61 & \textbf{0.53} & \underline{0.39} & 0.53 & \underline{0.26} & \underline{0.40} & 0.39 \\

\bottomrule
\end{tabular}
}%
}
\end{minipage}
\hfill
\begin{minipage}[t]{0.49\linewidth}
\centering
{\fontsize{7.5}{9.5}\selectfont
\setlength{\tabcolsep}{0.4mm}

\captionof{table}{
Performance comparison between FiRe-SFT and FiRe-GRPO on GenEval++.
}
\label{tab:sft_rl}

\resizebox{\linewidth}{!}{%
\begin{tabular}{lcccccccc}
\toprule
& \multicolumn{8}{c}{\textbf{GenEval++ $\uparrow$}} \\
\cmidrule{2-9}
\textbf{Method}
& Overall
& Color
& Count
& \makecell{Color/\\Count}
& \makecell{Color/\\Pos}
& \makecell{Pos/\\Count}
& \makecell{Pos/\\Size}
& \makecell{Multi-\\Count} \\
\midrule

Janus-Pro-7B
& 0.25 & 0.45 & 0.30 & 0.13 & 0.30 & 0.08 & 0.35 & 0.13 \\

\midrule

\textbf{FiRe-SFT} (1st)
& 0.35 & 0.58 & 0.28 & 0.24 & 0.49 & 0.11 & 0.39 & 0.36 \\

\textbf{FiRe-SFT} (2nd)
& 0.41 & 0.66 & 0.39 & 0.30 & 0.54 & 0.16 & 0.39 & 0.42 \\

\textbf{FiRe-SFT} (3rd)
& 0.44 & 0.68 & 0.44 & 0.34 & 0.56 & 0.20 & 0.39 & 0.46 \\

\midrule

\textbf{FiRe-GRPO} (1st)
& 0.38 & 0.63 & 0.33 & 0.28 & 0.55 & 0.18 & \underline{0.37} & 0.33 \\

\textbf{FiRe-GRPO} (2nd)
& \underline{0.46}
& \underline{0.75}
& \underline{0.44}
& \underline{0.38}
& \textbf{0.58}
& \underline{0.26}
& \textbf{0.38}
& \underline{0.46} \\

\textbf{FiRe-GRPO} (3rd)
& \textbf{0.49}
& \textbf{0.78}
& \textbf{0.49}
& \textbf{0.45}
& \underline{0.57}
& \textbf{0.28}
& 0.36
& \textbf{0.49} \\

\bottomrule
\end{tabular}
}%
}
\end{minipage}

\end{table*}

To better understand the sources of FiRe's improvements, we conduct a series of ablation studies and analyses. Notably, all FiRe-GRPO results in the ablation studies use the 300-step
checkpoint described in Appendix~\ref{app:training_details}. We first examine the effect of reasoning step granularity and the role of step-wise advantage estimation in FiRe-GRPO. We then compare FiRe-SFT and FiRe-GRPO to assess the contribution of FiRe-GRPO beyond supervised fine-tuning. Finally, we analyze generalization across model scales, computational overhead, and unedited-region preservation during iterative correction.

\paragraph{Effect of Reasoning Step Granularity. \quad}
To validate the effectiveness of FiRe's fine-grained reasoning process, we conduct two complementary ablations.
First, we compare the original multi-step reasoning path with a coarse single-step evaluation variant.
While FiRe decomposes reasoning into Steps 1--6, the coarse variant merges the intermediate evaluation Steps 2--4 into a single prompt-level assessment and directly generates feedback, thereby removing FiRe's fine-grained step-by-step alignment evaluation.
Second, we independently ablate the Tuple Decomposition (Step 3) and VQA (Step 4) steps to examine the necessity of individual components within the fine-grained reasoning process.
All variants use the same training images and identical SFT configuration, differing only in their intermediate reasoning supervision.
Details of the coarse reasoning variant are provided in Appendix~\ref{app:coarse_reasoning_ablation}.
As shown in Table~\ref{tab:ablation1}, FiRe's multi-step reasoning outperforms the coarse single-step variant across most GenEval++ categories, demonstrating the benefit of fine-grained reasoning.
Table~\ref{tab:fire_sft_ablation} further shows that removing either the Tuple Decomposition or VQA step reduces later-round performance, while removing both results in the largest degradation.
Together, these results demonstrate that FiRe's refinement gains arise from its fine-grained reasoning design rather than iterative editing alone.

\paragraph{Effect of Advantage Estimation in FiRe-GRPO. \quad}
For multi-step reasoning, a single outcome-level reward provides a sparse training signal~\citep{zhang2025process, wang2026discovering}: terminal rewards do not directly supervise intermediate steps, while aggregating step rewards mixes successful and unsuccessful steps into one trajectory-level signal.
FiRe-GRPO instead computes rewards and advantages for individual FiRe reasoning steps (w/ Stepwise).
We compare against w/ Terminal, which uses only the final image alignment reward, and w/ Aggregated, which uses the same step-specific rewards as w/ Stepwise but aggregates them into a single trajectory-level advantage shared across the reasoning steps.
As shown in Table~\ref{tab:ablation2}, w/ Stepwise performs best across most GenEval++ categories.
The improvement over w/ Terminal demonstrates the benefit of dense intermediate rewards, while the improvement over w/ Aggregated demonstrates the benefit of step-wise credit assignment.
All variants use the same Qwen3.5-35B-A3B~\citep{qwen35blog} reward evaluator.

\paragraph{Comparison between FiRe-SFT and FiRe-GRPO. \quad}
To assess the effectiveness of FiRe-GRPO, we compare FiRe-GRPO with its supervised fine-tuned counterpart, FiRe-SFT. As shown in Table~\ref{tab:sft_rl}, FiRe-SFT already outperforms the baselines on GenEval++, a complex compositional benchmark, indicating the effectiveness of the proposed fine-grained reasoning method itself. FiRe-GRPO further improves performance across all refinement iterations, isolating the additional benefit of reinforcement learning beyond supervised fine-tuning.

\paragraph{Generalizability Across Model Scales. \quad}
To verify scale generalizability, we apply the same SFT and GRPO training pipeline to Janus-Pro-1B. Consequently, FiRe consistently outperforms the base model (Appendix~\ref{app:1b_model}), demonstrating that our fine-grained reasoning framework remains effective across different backbone sizes.

\begin{figure}[t]
    \centering
    \includegraphics[width=\linewidth]{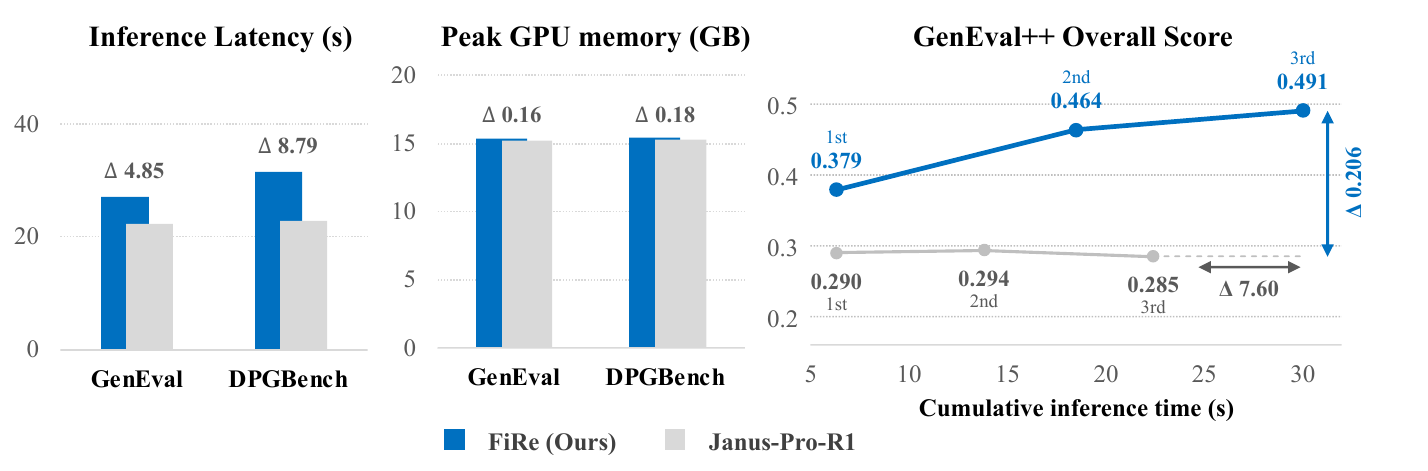}
    \caption{
    Efficiency and cost--performance comparison between FiRe and Janus-Pro-R1.
    Left: inference latency and peak GPU memory on GenEval and DPGBench~\citep{ghosh2023geneval,hu2024ella}.
    Right: GenEval++~\citep{ye2025echo} overall score as a function of cumulative inference time across three refinement rounds.
    FiRe achieves consistent performance gains with additional refinement, while Janus-Pro-R1 shows only a marginal second-round gain and degrades in the third round.
    }
    \label{fig:efficiency}
\end{figure}
\begin{table*}[t!]
\centering

\begin{minipage}[t]{0.62\linewidth}
\centering
\vspace{0pt}

\captionof{table}{
Human-preference and aesthetic quality across refinement rounds on GenEval++, measured by HPSv3 and LAION-Aesthetic-v2.
}
\label{tab:perceptual_quality}

{\fontsize{8.5}{10}\selectfont
\setlength{\tabcolsep}{1.7mm}

\resizebox{\linewidth}{!}{%
\begin{tabular}{lcccccc}
\toprule
& \multicolumn{3}{c}{\textbf{HPSv3 $\uparrow$}~\citep{ma2025hpsv3}}
& \multicolumn{3}{c}{\textbf{LAION-Aesthetic-v2 $\uparrow$}~\citep{schuhmann2022aesthetic}} \\
\cmidrule(lr){2-4}
\cmidrule(lr){5-7}
\textbf{Method}
& \textbf{1st}
& \textbf{2nd}
& \textbf{3rd}
& \textbf{1st}
& \textbf{2nd}
& \textbf{3rd} \\
\midrule
Janus-Pro~\citep{chen2025janus}
& 6.3600 & -- & --
& 5.512 & -- & -- \\

T2I-R1~\citep{jiang2025t2i}
& 7.3293 & -- & --
& 5.549 & -- & -- \\

Janus-Pro-R1~\citep{pan2025janus}
& 7.4088 & 6.7799 & 6.7188
& \textbf{5.685} & 5.569 & 5.543 \\

FiRe
& \textbf{8.7062}
& \textbf{8.5193}
& \textbf{8.5592}
& 5.608
& \textbf{5.602}
& \textbf{5.600} \\
\bottomrule
\end{tabular}
}%
}

\end{minipage}
\hfill
\begin{minipage}[t]{0.35\linewidth}
\centering
\vspace{0pt}

\captionof{table}{
Multimodal understanding performance before and after FiRe training.
}
\label{tab:understanding}

{\fontsize{8.5}{10}\selectfont
\setlength{\tabcolsep}{1.7mm}

\begin{tabular}{lcc}
\toprule
\textbf{Metric $\uparrow$}
& \textbf{Janus-Pro}
& \textbf{FiRe} \\
\midrule
MME-P~\citep{fu2026mme}
& 1502
& \textbf{1541} \\

POPE~\citep{li2023evaluating}
& \textbf{86.35}
& 85.97 \\

MMBench~\citep{liu2024mmbench}
& \textbf{78.30}
& 77.69 \\

SEED-Bench~\citep{li2023seed}
& \textbf{72.02}
& 71.31 \\

GQA~\citep{hudson2019gqa}
& \textbf{61.65}
& 60.10 \\
\bottomrule
\end{tabular}
}

\end{minipage}

\end{table*}

\paragraph{Computational Overhead Analysis. \quad}
We analyze FiRe's computational overhead in Figure~\ref{fig:efficiency} from both per-round efficiency and cumulative cost--performance perspectives. All inference latency measurements are obtained with SGLang~\citep{zheng2024sglang} on a single NVIDIA A100 GPU.
The left panel compares inference latency and peak GPU memory with Janus-Pro-R1~\citep{pan2025janus} on GenEval and DPGBench, showing that FiRe incurs additional latency from fine-grained reasoning while introducing only modest memory overhead.
The right panel shows the cost--performance trade-off across refinement rounds: FiRe consistently improves with additional inference time, whereas Janus-Pro-R1 yields only a marginal second-round gain and degrades in the third round.
Detailed latency breakdowns and scaling with prompt length and tuple count are provided in Appendix~\ref{app:inference_cost}.

\paragraph{Quality and Capability Preservation. \quad}
We further examine whether FiRe's iterative refinement and training compromise perceptual quality or multimodal understanding.
As shown in Table~\ref{tab:perceptual_quality}, FiRe maintains strong human-preference and aesthetic quality across refinement rounds, while consistently outperforming Janus-Pro-R1 on HPSv3~\citep{ma2025hpsv3}.
Table~\ref{tab:understanding} further shows that FiRe largely preserves the multimodal understanding capability of the Janus-Pro backbone, improving MME-P~\citep{fu2026mme} while showing only minor decreases on the remaining benchmarks.
We additionally evaluate preservation of unedited image regions in Appendix~\ref{app:unedit_preservation}.

\paragraph{Error Propagation, Failure, and Robustness Analysis. \quad}
We analyze FiRe's failure modes and robustness in Appendix~\ref{app:error_propagation}, including degradation across refinement rounds, step-level failure attribution, error recovery, and robustness to visually under-specified prompts.
FiRe exhibits fewer degradation cases than Janus-Pro-R1, explicit rechecking further improves refinement, and performance remains stable on visually under-specified prompts.

\paragraph{Fixed Initial Image Comparison. \quad}
To examine whether FiRe's gains simply come from better initial image
generation, we conduct a fixed-initial-image comparison in
Appendix~\ref{app:fixed_init}. Specifically, we fix the initial images to those
generated by Janus-Pro-R1 and compare subsequent refinement from the same
starting images. FiRe achieves higher overall performance on both GenEval and DPGBench, showing that the improvement is not solely due to initial image quality but also comes from fine-grained reasoning and localized correction.

\section{Related Works}

\paragraph{Unified MLLMs. \quad}
Recent unified MLLMs~\citep{chen2025janus, ma2025janusflow,ge2023making,ge2024seed,tong2025metamorph,xie2025show,ma2025unitok,wu2024vila,wang2024emu3} increasingly integrate image understanding and generation within a single architecture. 
By unifying perception and generation within a single model, these architectures enable self-refinement entirely within the model itself, extending the chain-of-thought, self-reflection techniques extensively studied in image understanding~\citep{zhang2025r1-vl, xu2024r1-llava-cot, thawakar2025r1-llamaV-o1, yao2024mulberry, zhang2024vlm-cot, huang2025visionr1}. 
This reasoning capability can also be leveraged to enhance image generation, allowing the model's own deliberative reasoning developed for perception to extend to generation as well.

\paragraph{Multimodal Reasoning for Text-To-Image Generation. \quad}
Recent work on MLLM-based image generation via intermediate reasoning broadly falls into two paradigms: prompt augmentation~\citep{jiang2025t2i, zhang2025reasongen, duan2025got}, which enriches input prompts but lacks post-generation correction, and iterative self-reflection~\citep{pan2025janus, qin2025uni,ye2026understanding}. Both paradigms rely on coarse VQA judgments and generate feedback from this holistic assessment.
FiRe instead performs six-step fine-grained reasoning.
This enables FiRe to address fine-grained prompt--image mismatches that prompt augmentation and coarse self-reflection fail to capture.

\paragraph{Credit Assignment in GRPO. \quad}
Recently, Group Relative Policy Optimization (GRPO)~\citep{shao2024deepseekmath, guo2025deepseek} has emerged as an efficient policy optimization algorithm for LLM and MLLM training that estimates advantages from group-relative final-answer signals without a separate value function.
However, in long Chain-of-Thought reasoning, GRPO's trajectory-level advantage yields coarse credit assignment, treating every token in a trajectory as equally responsible for the final outcome. To address this, recent works have investigated fine-grained credit assignment within GRPO—for example, step-wise advantages for multimodal reasoning~\citep{zhang2025r1-vl}, segment-level advantages for Chain-of-Thought reasoning~\citep{guo2025segment, kazemnejad2024vineppo}, and turn-level advantages for multi-turn tool use~\citep{wei2025reinforcing}.
Following recent step-level extensions of GRPO such as GiGPO~\citep{feng2025group}, FiRe-GRPO likewise enables fine-grained credit assignment by computing a separate group-relative advantage at each step rather than broadcasting a single trajectory-level signal.

\section{Limitations and Conclusion}

We introduced FiRe, a fine-grained multimodal reasoning method for text-to-image generation that structures iterative refinement into a six-step reasoning process, and FiRe-GRPO, which improves FiRe beyond supervised imitation by optimizing step-level reasoning and correction.
Experiments show that FiRe improves prompt adherence across compositional benchmarks, especially on complex prompts involving attributes, counts, and spatial relations.
As with other iterative image refinement methods, FiRe naturally depends on the reliability of intermediate reasoning outputs across refinement rounds; FiRe-GRPO mitigates this error accumulation problem by optimizing these steps with step-level rewards, and our error propagation analysis in Appendix~\ref{app:error_propagation} shows fewer degradation cases than Janus-Pro-R1~\citep{pan2025janus}.
FiRe may also be limited in explicitly verifying visually under-specified semantics, while our directly synthesized compositional training data rely on a single generation/editing model family and FiRe-GRPO uses a single VLM-based reward evaluator.
Future work can further improve robustness through reasoning verification, synthetic training data generated from diverse image generation and editing models, reward evaluation with multiple evaluators, and explicit step-level branching across diverging trajectories---as in GiGPO~\citep{feng2025group}---to broaden exploration beyond a single rollout, complementing FiRe's per-step credit assignment.

\clearpage

\begin{ack}
This work was supported by KT (Korea Telecom)-Korea University AICT R\&D Center. This work was also supported by the National Research Foundation of Korea (NRF) grant funded by the Korean government (MSIT) (No.~RS-2026-25613293, Project Title: Physics- and Mathematics-Informed Causal AI Models for Robust Analysis and Prediction under Geometry, Boundary-Condition, Parameter, and Multi-Environment Shifts, 20\%), (No.~RS-2024-00353007, 10\%), and (No.~RS-2024-00410082, AI Foundation Model for Solving Combinatorial Optimization Problems and Its Applications to Various Industrial Problems, 10\%), and was supported by the Institute of Information \& Communications Technology Planning \& Evaluation (IITP) grant funded by the Korea government (MSIT) (No.~RS-2022-II220612, Project Title: Geometric and Physical Commonsense Reasoning based Behavior Intelligence for Embodied AI, 10\%), (IITP-2026-RS-2025-02304828, artificial intelligence star fellowship support program to nurture the best talents, 20\%), (IITP-2026-RS-2024-00436857, ITRC (Information Technology Research Center), 10\%), and (No.~RS-2026-25507543, Development of AI Co-Scientist based on Scientific Causal World Model, 20\%).
\end{ack}

{
\small
\bibliographystyle{unsrtnat}
\bibliography{references}
}


\appendix

\clearpage
\section{Broader Impacts}
This work introduces FiRe, designed to enhance the semantic alignment and quality of text-to-image generation through iterative reasoning. 

\paragraph{Positive Societal Impacts} Our method provides a robust mechanism for ensuring that generated visual content strictly adheres to complex human instructions. This has significant potential in creative industries, enabling users to perform fine-grained visual edits via natural language, thereby lowering the barrier for high-quality digital content creation. Furthermore, the underlying fine-grained verification logic contributes to the broader field of AI safety and reliability by making the generative process more transparent and controllable.

\paragraph{Ethical Considerations} As with any advanced image generation and editing technology, there is a risk of potential misuse for creating deceptive content or misinformation. We emphasize that this technology should be used in conjunction with robust authentication and deepfake detection methods. Additionally, since our method relies on large-scale pre-trained models (LLMs and VLMs), it may inadvertently reflect or amplify biases present in its training data. We encourage users and developers to remain vigilant regarding algorithmic fairness and the ethical implications of synthesized visual media.

\section{Detailed Input Prompts for FiRe}
\label{app:fire_prompts}

This section provides the input prompts used in FiRe. 

\paragraph{Step~1: Initial Text-to-Image Generation.}
For initial image generation, FiRe takes the original text prompt $p$ as input
and generates an initial image $I$. The input consists of the user prompt
followed by the image generation start token.

\begin{tcblisting}{
    listing only, breakable, colback=gray!5, colframe=gray!75,
    fonttitle=\bfseries, fontupper=\scriptsize,
    label={app:prompt_step1},
    title={Input Prompt: Step~1 Initial Text-to-Image Generation},
    listing options={basicstyle=\ttfamily\scriptsize, breaklines=true}
}
<|User|>
{p}

<|Assistant|>
{image_start_tag}
\end{tcblisting}

\paragraph{Steps~2--5: Fine-grained Reasoning and Feedback Generation.}
For fine-grained reasoning, FiRe conditions on the original prompt $p$ and the
current image $I$, and uses a single input prompt to generate Steps~2--5 as one
autoregressive reasoning trace. Although the external input to this prompt is
$(p,I)$, the effective context is expanded as the model generates each
intermediate output. Specifically, Step~2 generates the summarized prompt
$S$ from $(p,I)$; Step~3 generates the tuple decomposition $D$ conditioned on
$(p,I,S)$; Step~4 generates the tuple-level verification results $V$
conditioned on $(p,I,S,D)$; and Step~5 generates the feedback $F$ conditioned on
$(p,I,S,D,V)$. This construction allows later reasoning steps to directly use
the outputs of earlier steps while requiring only one input prompt for the full
reasoning trace.

\begin{tcblisting}{
    listing only, breakable, colback=gray!5, colframe=gray!75,
    fonttitle=\bfseries, fontupper=\scriptsize,
    label={app:prompt_steps2_5},
    title={Input Prompt: Steps~2--5 Fine-grained Reasoning and Feedback},
    listing options={basicstyle=\ttfamily\scriptsize, breaklines=true}
}
<|User|>
{p}

<|Assistant|>
{image_start_tag}{I}{image_end_tag}

First, summarize the input prompt by keeping only explicitly stated visual facts
\end{tcblisting}

\paragraph{Step~6: Localized Image Correction.}
For localized correction, FiRe takes the current image $I$, the original prompt
$p$, and the feedback $F$ as input, and generates a corrected image $I'$. The
editing prompt instructs the model to apply all feedback instructions while
preserving visual content that does not need to be changed.

\begin{tcblisting}{
    listing only, breakable, colback=gray!5, colframe=gray!75,
    fonttitle=\bfseries, fontupper=\scriptsize,
    label={app:prompt_step6},
    title={Input Prompt: Step~6 Localized Image Correction},
    listing options={basicstyle=\ttfamily\scriptsize, breaklines=true}
}
<|User|>
{image_start_tag}{I}{image_end_tag}

Please edit the image as instructed.
The FEEDBACK will be given as multiple correction requests.
You MUST apply all requested corrections and produce a final image reflecting all changes.
Preserve the background, objects, style, spatial layout, and composition of the source image unless explicitly asked to change them.

INPUT_PROMPT: {p}

FEEDBACK:
{F}

<|Assistant|>
{image_start_tag}
\end{tcblisting}

\section{Detailed FiRe Inference Algorithm}
\label{app:fire_inference}

Given a text prompt $p$, FiRe aims to generate an image $I$ that satisfies the fine-grained visual conditions in $p$, such as object presence, attributes, counts, and spatial relations. FiRe achieves this through six fine-grained reasoning steps, iteratively refining the image until all visual requirements are satisfied or a maximum number of attempts $H$ is reached. The overall inference algorithm is presented in \Cref{alg:fire_inference}.

\begin{algorithm}[!htbp]
\caption{FiRe Inference (input prompt $p$)}
\label{alg:fire_inference}
\begin{algorithmic}[1]
\Require Prompt $p$, policy $\pi_\theta$, maximum number of correction attempts $H$
\Ensure Final image $I$ aligned with the input prompt $p$
\State $I \gets \pi_\theta(p)$ \Comment{Step 1: initial T2I generation}
\For{$k = 1, \ldots, H$}
    \State $S \gets \pi_\theta(p, I)$ \Comment{Step 2: prompt summarization}
    \State $D = \{\tau_i\}_{i=1}^{Q} \gets \pi_\theta(p, I, S)$
        \Comment{Step 3: tuple decomposition}
    \State $V = \{(r_i, y_i)\}_{i=1}^{Q} \gets \pi_\theta(p, I, S, D)$
        \Comment{Step 4: tuple-level VQA, $v_i = (r_i, y_i)$}
    \State $F \gets \pi_\theta(p, I, S, D, V)$
        \Comment{Step 5: fine-grained feedback}
    \If{$F = \texttt{no-edit}$}
        \State \textbf{break} \Comment{all tuples satisfied}
    \EndIf
    \State $I' \gets \pi_\theta(p, I, F)$
        \Comment{Step 6: localized correction}
    \State $I \gets I'$
        \Comment{use corrected image as input to next attempt}
\EndFor
\State \Return $I$
\end{algorithmic}
\end{algorithm}

\section{Data Preparation Details}
\label{app:data_preparation_details}

We construct supervised FiRe trajectories from two data sources: an edit dataset of 140K samples derived from FocusDiff~\citep{pan2025focusdiff} and a composition-specific synthetic dataset of 60K samples, yielding 200K image-text pairs for SFT in total. The synthetic samples are generated with Qwen-Image and Qwen-Image-Edit~\citep{wu2025qwen} from prompts adapted from~\citep{liu2025flow, zhang2025reasongen}. For FiRe-GRPO, we additionally incorporate complex and long prompts from~\citep{liu2024improving, ye2025echo} to expose the model to more challenging compositional reasoning scenarios.

Throughout this section, we use Qwen3-Next-80B-A3B~\citep{yang2025qwen3} as the text annotation model, denoted $\mathcal{M}_{\mathrm{text}}$, for prompt summarization, tuple decomposition, question generation, and feedback generation, and Qwen3-VL-32B~\citep{bai2025qwen3vltechnicalreport} as the VQA model, denoted $\mathcal{M}_{\mathrm{vqa}}$, for tuple-level VQA.

\subsection{Edit Dataset}
\label{app:edit_dataset}

We construct supervised FiRe trajectories from the FocusDiff dataset
\citep{pan2025focusdiff}, which contains paired images before and after editing,
denoted by $I_{\mathrm{pre}}$ and $I_{\mathrm{post}}$, together with their
corresponding captions $p_{\mathrm{pre}}$ and $p_{\mathrm{post}}$. Our goal is
to convert each edit pair into FiRe-style supervision consisting of a current
image $I$, summarized prompt $S$, tuple decomposition $D$, tuple-level VQA
results $V$, feedback $F$, and, when correction is required, a corrected image
$I'$.

\paragraph{Prompt summarization, tuple decomposition, and VQA.}
We first set the post-edit caption $p_{\mathrm{post}}$ as the target prompt
$p$. The text annotation model summarizes $p$ into $S$ and decomposes $S$ into
a tuple set:
\begin{equation}
S=\mathcal{M}_{\mathrm{text}}(p),
\qquad
D=\{\tau_i\}_{i=1}^{Q}
=
\mathcal{M}_{\mathrm{text}}(p,S).
\end{equation}
Each tuple $\tau_i$ represents an atomic visual element to be checked in the
image. The text annotation model then converts each tuple $\tau_i$ into a
verification question:
\begin{equation}
q_i=\mathcal{M}_{\mathrm{text}}(\tau_i),
\qquad
i=1,\ldots,Q.
\end{equation}
Given each question, the VQA model evaluates both the pre-edit and post-edit
images:
\begin{equation}
v_i(I)
=
(r_i(I),y_i(I))
=
\mathcal{M}_{\mathrm{vqa}}(I,q_i),
\qquad
I\in\{I_{\mathrm{pre}},I_{\mathrm{post}}\},
\end{equation}
where $r_i(I)$ is an image-grounded rationale and
$y_i(I)\in\{\texttt{yes},\texttt{no}\}$ indicates whether image $I$ satisfies
tuple $\tau_i$. The full tuple-level VQA results for image $I$ are denoted by
\begin{equation}
V(I)=\{v_i(I)\}_{i=1}^{Q}.
\end{equation}

\paragraph{Filtering and role assignment.}
For the target prompt $p=p_{\mathrm{post}}$, we keep an edit pair only if the
post-edit image satisfies all tuples while the pre-edit image violates at least
one tuple:
\begin{equation}
y_i(I_{\mathrm{post}})=\texttt{yes}\ \ \forall i,
\qquad
\exists i \text{ such that } y_i(I_{\mathrm{pre}})=\texttt{no}.
\end{equation}
We then assign the post-edit image as the prompt-aligned image and the pre-edit
image as the prompt-misaligned image:
\begin{equation}
I_{\mathrm{align}}=I_{\mathrm{post}},
\qquad
I_{\mathrm{mis}}=I_{\mathrm{pre}}.
\end{equation}
Samples that do not meet these conditions are discarded.

\paragraph{Trajectory construction.}
For the prompt-misaligned image $I_{\mathrm{mis}}$, we construct a corrective
trajectory. Its tuple-level VQA result is
\begin{equation}
V_{\mathrm{mis}}=V(I_{\mathrm{mis}}).
\end{equation}
The text annotation model generates corrective feedback from the target prompt,
tuple set, and tuple-level VQA results:
\begin{equation}
F_{\mathrm{mis}}
=
\mathcal{M}_{\mathrm{text}}(p,D,V_{\mathrm{mis}}),
\end{equation}
where the feedback targets tuples with
$y_i(I_{\mathrm{mis}})=\texttt{no}$. The prompt-aligned image is used as the
corrected image target, i.e., $I'=I_{\mathrm{align}}$. This yields the
supervised corrective trajectory
\begin{equation}
\tau_{\mathrm{sft}}^{\mathrm{mis}}
=
(I_{\mathrm{mis}},S,D,V_{\mathrm{mis}},F_{\mathrm{mis}},I_{\mathrm{align}}).
\end{equation}

For the prompt-aligned image $I_{\mathrm{align}}$, we construct a no-edit
trajectory:
\begin{equation}
V_{\mathrm{align}}=V(I_{\mathrm{align}}),
\qquad
F_{\mathrm{align}}=\texttt{no-edit}.
\end{equation}
This yields the supervised no-edit trajectory
\begin{equation}
\tau_{\mathrm{sft}}^{\mathrm{align}}
=
(I_{\mathrm{align}},S,D,V_{\mathrm{align}},F_{\mathrm{align}}).
\end{equation}

\paragraph{Bidirectional expansion.}
To increase data coverage, we repeat the same construction in the reverse direction by setting the pre-edit caption $p_{\mathrm{pre}}$ as the target prompt $p$. In this case, $S$, $D$, and the verification questions are regenerated from $p_{\mathrm{pre}}$, and the roles are reversed: $I_{\mathrm{pre}}$ is treated as the prompt-aligned image if it satisfies all tuples derived from $p_{\mathrm{pre}}$, while $I_{\mathrm{post}}$ is treated as the prompt-misaligned image if it violates at least one tuple. We apply the same filtering and trajectory construction procedure to obtain additional corrective and no-edit trajectories.

\subsection{Composition-specific Synthetic Dataset}
\label{app:synthetic_dataset}

To enhance compositional reasoning, we construct a composition-specific
synthetic dataset using prompts that follow the styles of established
benchmarks such as GenEval~\citep{ghosh2023geneval} and DPGBench~\citep{hu2024ella}. We use
Qwen-Image~\citep{wu2025qwen} as the image generation model, denoted by
$\mathcal{M}_{\mathrm{gen}}$, and Qwen-Image-Edit~\citep{wu2025qwen} as the
image editing model, denoted by $\mathcal{M}_{\mathrm{edit}}$. To avoid data
contamination, we strictly exclude all prompts that appear in the final
evaluation sets of these benchmarks from the synthetic training pipeline.

\paragraph{Source prompt generation and verification.}
We first sample a source prompt $p_{\mathrm{src}}$ and generate an initial
image:
\begin{equation}
I_{\mathrm{init}}
=
\mathcal{M}_{\mathrm{gen}}(p_{\mathrm{src}}).
\end{equation}
To ensure that $I_{\mathrm{init}}$ is a reliable starting point, we verify its
alignment with $p_{\mathrm{src}}$. The text annotation model summarizes
$p_{\mathrm{src}}$ and decomposes it into tuples:
\begin{equation}
S_{\mathrm{src}}
=
\mathcal{M}_{\mathrm{text}}(p_{\mathrm{src}}),
\qquad
D_{\mathrm{src}}
=
\{\tau_i^{\mathrm{src}}\}_{i=1}^{Q_{\mathrm{src}}}
=
\mathcal{M}_{\mathrm{text}}(p_{\mathrm{src}},S_{\mathrm{src}}).
\end{equation}
It then converts each tuple into a verification question:
\begin{equation}
q_i^{\mathrm{src}}
=
\mathcal{M}_{\mathrm{text}}(\tau_i^{\mathrm{src}}),
\qquad
i=1,\ldots,Q_{\mathrm{src}}.
\end{equation}
The VQA model evaluates the generated image with respect to each question:
\begin{equation}
v_i^{\mathrm{src}}(I_{\mathrm{init}})
=
(r_i^{\mathrm{src}}(I_{\mathrm{init}}),
y_i^{\mathrm{src}}(I_{\mathrm{init}}))
=
\mathcal{M}_{\mathrm{vqa}}(I_{\mathrm{init}},q_i^{\mathrm{src}}),
\end{equation}
where $y_i^{\mathrm{src}}\in\{\texttt{yes},\texttt{no}\}$. We keep the sample
only if $I_{\mathrm{init}}$ satisfies all tuples of $p_{\mathrm{src}}$:
\begin{equation}
y_i^{\mathrm{src}}(I_{\mathrm{init}})=\texttt{yes}
\quad
\forall i.
\end{equation}

\paragraph{Target prompt construction and cross-prompt verification.}
Next, we construct a target prompt $p_{\mathrm{tgt}}$ by modifying
$p_{\mathrm{src}}$ with contrasting compositional attributes, counts, or spatial
relations. The text annotation model summarizes $p_{\mathrm{tgt}}$ and
decomposes it into target tuples:
\begin{equation}
S_{\mathrm{tgt}}
=
\mathcal{M}_{\mathrm{text}}(p_{\mathrm{tgt}}),
\qquad
D_{\mathrm{tgt}}
=
\{\tau_i^{\mathrm{tgt}}\}_{i=1}^{Q_{\mathrm{tgt}}}
=
\mathcal{M}_{\mathrm{text}}(p_{\mathrm{tgt}},S_{\mathrm{tgt}}).
\end{equation}
Each target tuple is converted into a verification question:
\begin{equation}
q_i^{\mathrm{tgt}}
=
\mathcal{M}_{\mathrm{text}}(\tau_i^{\mathrm{tgt}}),
\qquad
i=1,\ldots,Q_{\mathrm{tgt}}.
\end{equation}
We then evaluate the initial image $I_{\mathrm{init}}$ against the target
prompt using the VQA model:
\begin{equation}
v_i^{\mathrm{tgt}}(I_{\mathrm{init}})
=
(r_i^{\mathrm{tgt}}(I_{\mathrm{init}}),
y_i^{\mathrm{tgt}}(I_{\mathrm{init}}))
=
\mathcal{M}_{\mathrm{vqa}}(I_{\mathrm{init}},q_i^{\mathrm{tgt}}),
\qquad
i=1,\ldots,Q_{\mathrm{tgt}}.
\end{equation}
The resulting tuple-level VQA set is
\begin{equation}
V_{\mathrm{init}}^{\mathrm{tgt}}
=
\{v_i^{\mathrm{tgt}}(I_{\mathrm{init}})\}_{i=1}^{Q_{\mathrm{tgt}}}.
\end{equation}

\paragraph{Selection and role assignment.}
We retain the sample only if $I_{\mathrm{init}}$ is misaligned with respect to
the target prompt $p_{\mathrm{tgt}}$:
\begin{equation}
\exists i
\text{ such that }
y_i^{\mathrm{tgt}}(I_{\mathrm{init}})=\texttt{no}.
\end{equation}
If $I_{\mathrm{init}}$ already satisfies all tuples of $p_{\mathrm{tgt}}$, the
sample is discarded because it provides no corrective signal. For retained
samples, we assign
\begin{equation}
I_{\mathrm{mis}}=I_{\mathrm{init}},
\qquad
V_{\mathrm{mis}}=V_{\mathrm{init}}^{\mathrm{tgt}},
\end{equation}
with the target annotations
\begin{equation}
p=p_{\mathrm{tgt}},
\qquad
S=S_{\mathrm{tgt}},
\qquad
D=D_{\mathrm{tgt}}.
\end{equation}

\paragraph{Feedback generation, editing, and finalization.}
From the target prompt $p$, tuple set $D$, and tuple-level VQA results
$V_{\mathrm{mis}}$, the text annotation model generates corrective feedback:
\begin{equation}
F_{\mathrm{mis}}
=
\mathcal{M}_{\mathrm{text}}(p,D,V_{\mathrm{mis}}).
\end{equation}
The editing model then produces a refined image conditioned on the misaligned
image and feedback:
\begin{equation}
I_{\mathrm{edit}}
=
\mathcal{M}_{\mathrm{edit}}(p,I_{\mathrm{mis}},F_{\mathrm{mis}}).
\end{equation}
We evaluate $I_{\mathrm{edit}}$ against the target prompt using the target
verification questions:
\begin{equation}
v_i^{\mathrm{tgt}}(I_{\mathrm{edit}})
=
(r_i^{\mathrm{tgt}}(I_{\mathrm{edit}}),
y_i^{\mathrm{tgt}}(I_{\mathrm{edit}}))
=
\mathcal{M}_{\mathrm{vqa}}(I_{\mathrm{edit}},q_i^{\mathrm{tgt}}),
\qquad
i=1,\ldots,Q_{\mathrm{tgt}}.
\end{equation}
If the edited image satisfies all target tuples,
\begin{equation}
y_i^{\mathrm{tgt}}(I_{\mathrm{edit}})=\texttt{yes}
\quad
\forall i,
\end{equation}
we finalize it as the prompt-aligned image:
\begin{equation}
I_{\mathrm{align}}=I_{\mathrm{edit}}.
\end{equation}
This yields the supervised corrective trajectory
\begin{equation}
\tau_{\mathrm{sft}}^{\mathrm{mis}}
=
(I_{\mathrm{mis}},S,D,V_{\mathrm{mis}},F_{\mathrm{mis}},I_{\mathrm{align}}).
\end{equation}

We additionally construct the corresponding no-edit trajectory for the aligned
image:
\begin{equation}
V_{\mathrm{align}}
=
\{v_i^{\mathrm{tgt}}(I_{\mathrm{align}})\}_{i=1}^{Q_{\mathrm{tgt}}},
\qquad
F_{\mathrm{align}}=\texttt{no-edit},
\end{equation}
\begin{equation}
\tau_{\mathrm{sft}}^{\mathrm{align}}
=
(I_{\mathrm{align}},S,D,V_{\mathrm{align}},F_{\mathrm{align}}).
\end{equation}

\section{Step-specific Reward Design}
\label{app:reward}

This section provides the detailed reward computation used in FiRe-GRPO.
We follow the notation in the main text: given a prompt $p$, FiRe produces
the current image $I$, summarized prompt $S$, tuple decomposition
$D=\{\tau_i\}_{i=1}^{Q}$, tuple-level VQA results
$V=\{v_i\}_{i=1}^{Q}$ with $v_i=(r_i,y_i)$, feedback $F$, and corrected image
$I'$. We use Qwen-3.5-35B-A3B~\citep{qwen35blog} as the VLM-based evaluator,
denoted by $\mathcal{E}_{\mathrm{vlm}}$, for all reward computations. Each
evaluation call is conditioned on a step-specific system prompt $\psi_k$ that
specifies the scoring criteria and output format for Step~$k$. The full
system prompts are provided in Appendix~\ref{app:system_prompts}.

\paragraph{Step~1 reward: Initial image generation.}
Step~1 generates the initial image $I$. Since Step~1 produces an image rather
than a structured textual output, we do not apply a format gate and set
$g_1(I)=1$. We evaluate $I$ using a fine-grained prompt--image alignment score.
This score follows the same high-level structure as FiRe's reasoning process:
the evaluator summarizes the prompt, decomposes it into semantic tuples, and
then verifies the image at a fine-grained level. Unlike FiRe's VQA step, which
directly checks each tuple through tuple-level VQA, the reward evaluator follows
prompt-decomposition-based image evaluation protocols
\citep{hu2023tifa,cho2023davidsonian,hu2024ella} by converting each tuple into
a VQA question.

Concretely, given a prompt $p$, the evaluator $\mathcal{E}_{\mathrm{vlm}}$
follows the prompt summarization and tuple decomposition process used in FiRe to
obtain an evaluator-side tuple set:
\begin{equation}
D^{\star}(p)=\{\tau_i^{\star}\}_{i=1}^{Q_p}.
\end{equation}
Each tuple $\tau_i^{\star}$ is then converted into a corresponding VQA question
$q_i^{\star}$:
\begin{equation}
\mathcal{Q}^{\star}(p)=\{q_i^{\star}\}_{i=1}^{Q_p}.
\end{equation}
For example, the tuple
$\tau_i^{\star}=\texttt{attribute-color(oven, pink)}$
can be converted into
$q_i^{\star}=\text{``Is the oven pink?''}$.
For each question, the evaluator checks whether the image satisfies the
corresponding visual condition:
\begin{equation}
b_i(I,p)
=
\mathcal{E}_{\mathrm{vlm}}(I,q_i^\star;\psi_1),
\qquad
b_i(I,p)\in\{0,1\}.
\end{equation}
The fine-grained image alignment score $z$ is the average satisfaction score over
all questions:
\begin{equation}
z_{\mathrm{img}}(I,p)
=
\frac{1}{Q_p}
\sum_{i=1}^{Q_p} b_i(I,p),
\qquad
z_{\mathrm{img}}(I,p)\in[0,1].
\end{equation}
The Step~1 reward criterion is therefore
\begin{equation}
R_1(s_h,a_h)
=
g_1(I)\,z_{\mathrm{img}}(I,p)
=
z_{\mathrm{img}}(I,p).
\end{equation}

\paragraph{Step~2 reward: Prompt summarization.}
Step~2 generates the summarized prompt $S$. We evaluate whether $S$ retains
concrete visual details from the prompt $p$ that can be explicitly checked in
the image, such as objects, attributes, counts, spatial relations, and text,
while removing subjective or non-verifiable descriptions. The evaluator returns:
\begin{equation}
\tilde{z}_2
=
\mathcal{E}_{\mathrm{vlm}}(p,I,S;\psi_2),
\qquad
\tilde{z}_2\in[0,2].
\end{equation}
Since the summarized prompt must follow the required format to be used by
subsequent FiRe steps, we apply a format gate $g_2(S)\in\{0,1\}$. The Step~2
reward criterion is
\begin{equation}
R_2(s_h,a_h)
=
g_2(S)\frac{\tilde{z}_2}{2},
\qquad
R_2(s_h,a_h)\in[0,1].
\end{equation}

\paragraph{Step~3 reward: Tuple decomposition.}
Step~3 generates the tuple decomposition $D=\{\tau_i\}_{i=1}^{Q}$ from the
summarized prompt $S$. We evaluate whether $D$ faithfully decomposes the
visual content in $S$ into atomic visual elements, without omitting required
details or introducing hallucinated tuples. The evaluator returns:
\begin{equation}
\tilde{z}_3
=
\mathcal{E}_{\mathrm{vlm}}(p,S,D;\psi_3),
\qquad
\tilde{z}_3\in[0,2].
\end{equation}
We apply a format gate $g_3(D)\in\{0,1\}$ to ensure that the tuple set follows
the required structure. The Step~3 reward criterion is
\begin{equation}
R_3(s_h,a_h)
=
g_3(D)\frac{\tilde{z}_3}{2},
\qquad
R_3(s_h,a_h)\in[0,1].
\end{equation}

\paragraph{Step~4 reward: Tuple VQA.}
Step~4 generates tuple-level VQA results
$V=\{v_i\}_{i=1}^{Q}$, where $v_i=(r_i,y_i)$. We evaluate whether each result
contains an image-grounded rationale $r_i$ and an accurate binary judgment
$y_i\in\{\texttt{yes},\texttt{no}\}$ for the corresponding tuple $\tau_i$.
The evaluator returns
\begin{equation}
\tilde{z}_4
=
\mathcal{E}_{\mathrm{vlm}}(p,I,D,V;\psi_4),
\qquad
\tilde{z}_4\in[0,2].
\end{equation}
We apply a format gate $g_4(V)\in\{0,1\}$ to ensure that each VQA result
contains the required rationale and judgment fields. The Step~4 reward
criterion is
\begin{equation}
R_4(s_h,a_h)
=
g_4(V)\frac{\tilde{z}_4}{2},
\qquad
R_4(s_h,a_h)\in[0,1].
\end{equation}

\paragraph{Step~5 reward: Fine-grained feedback generation.}
Step~5 generates feedback $F$ from the tuple-level VQA results $V$. We evaluate
whether $F$ makes the correct decision between \texttt{no-edit} and correction
feedback, and whether it provides actionable guidance for tuples judged as
\texttt{no}. The evaluator returns
\begin{equation}
\tilde{z}_5
=
\mathcal{E}_{\mathrm{vlm}}(D,V,F;\psi_5),
\qquad
\tilde{z}_5\in[0,2].
\end{equation}
We apply a format gate $g_5(F)\in\{0,1\}$ to ensure that the feedback follows
the required structure. The Step~5 reward criterion is
\begin{equation}
R_5(s_h,a_h)
=
g_5(F)\frac{\tilde{z}_5}{2},
\qquad
R_5(s_h,a_h)\in[0,1].
\end{equation}

\paragraph{Step~6 reward: Localized image correction.}
Step~6 is performed only when $F\neq\texttt{no-edit}$. If
$F=\texttt{no-edit}$, the correction process terminates and no Step~6 action
or reward is assigned. Otherwise, FiRe generates a corrected image $I'$. Since
Step~6 produces an image rather than a structured textual output, we do not
apply a format gate and set $g_6(I')=1$. The corrected image is evaluated using
both prompt--image alignment and feedback instruction following.

The prompt--image alignment term uses the same fine-grained image alignment
score defined for Step~1:
\begin{equation}
z_{\mathrm{img}}(I',p)
=
\frac{1}{Q_p}
\sum_{i=1}^{Q_p}
\mathbf{1}\!\left[
\mathcal{E}_{\mathrm{vlm}}(I',q_i^\star;\psi_1)=\texttt{yes}
\right].
\end{equation}
The instruction-following score evaluates whether the edit from the current
image $I$ to the corrected image $I'$ follows the feedback $F$:
\begin{equation}
z_{\mathrm{if}}(I,F,I')
=
\frac{
\mathcal{E}_{\mathrm{vlm}}(I,F,I';\psi_{\mathrm{if}})
}{2},
\qquad
z_{\mathrm{if}}(I,F,I')\in[0,1].
\end{equation}
The Step~6 reward criterion is defined as the geometric mean of these two
scores:
\begin{equation}
R_6(s_h,a_h)
=
g_6(I')
\left(
z_{\mathrm{img}}(I',p)\cdot
z_{\mathrm{if}}(I,F,I')
\right)^{1/2}
=
\left(
z_{\mathrm{img}}(I',p)\cdot
z_{\mathrm{if}}(I,F,I')
\right)^{1/2}.
\end{equation}

\section{Benchmark Details}
\label{app:benchmark_details}

We evaluate on three benchmarks-GenEval, GenEval++, and DPGBench-covering a spectrum from short object-centric prompts to long, semantically dense prompts. For GenEval and GenEval++, we strictly follow the official evaluation protocols. For DPGBench, we observed that the official evaluation code produces inconsistent scores under multi-GPU configurations; we therefore use a corrected variant of the script and re-evaluate all baselines under this same corrected setup to ensure fair comparison. Detailed descriptions of each benchmark are provided below.

\subsection{GenEval}
GenEval~\citep{ghosh2023geneval} evaluates object-focused image generation using 550 text prompts. The prompts are systematically structured into six primary categories that measure fine-grained compositional adherence: \textbf{Single Object}, \textbf{Two Objects}, \textbf{Counting}, \textbf{Colors}, \textbf{Position}, and \textbf{Color Attribute Binding}.

\subsection{GenEval++}
GenEval++~\citep{ye2025echo} is a more challenging extension of GenEval for evaluating instruction-following fidelity in text-to-image generation. While GenEval focuses on relatively simple object-centric prompts, GenEval++ introduces more complex compositional instructions involving richer attribute combinations, counting, spatial relations, object positions, colors, and sizes. It consists of 280 high-complexity prompts across seven task types, with 40 prompts per task type. Compared with the original GenEval, GenEval++ places greater emphasis on difficult attribute binding and multi-condition satisfaction, making it less prone to metric saturation.

\subsection{DPGBench}
DPGBench~\citep{hu2024ella} evaluates compositional image generation with 1{,}065 \textbf{long} and \textbf{semantically rich} text prompts. The benchmark is built from COCO, PartiPrompts, DSG-1k, and Objects365, and its prompts are automatically extended by GPT-4 to describe multiple objects with diverse attributes and relationships. Compared to prior benchmarks such as GenEval~\citep{ghosh2023geneval}, DPGBench contains far more distinct nouns (4{,}286) and significantly longer prompts, enabling \textit{fine-grained} automatic evaluation of \textit{dense prompt} following via an MLLM-based VQA pipeline.

\section{FiRe Implementation Details}
\label{app:implementation_details}

\subsection{Training Details}
\label{app:training_details}

\paragraph{Supervised fine-tuning.}
We first perform supervised fine-tuning (SFT) to equip the base Janus-Pro models with FiRe's fine-grained multimodal reasoning ability. We train two model variants: Janus-Pro-7B and Janus-Pro-1B. For Janus-Pro-7B, we use eight NVIDIA A100 GPUs with DeepSpeed ZeRO-2 for memory-efficient training, performing full fine-tuning on the aligner, generation aligner, LLM backbone, and image head for 14K steps with a global batch size of 128, amounting to approximately 768 GPU hours. For Janus-Pro-1B, we use eight NVIDIA H100 GPUs with Distributed Data Parallel (DDP) and fine-tune the same set of components for 15K steps, amounting to approximately 120 GPU hours. The SFT hyperparameters are summarized in Table~\ref{tab:sft_hyperparam} and Table~\ref{tab:sft_optimizer}.

\begin{table}[!htbp]
  \centering
    \caption{SFT training setup for Janus-Pro models.}
  \begin{small}
    \setlength{\tabcolsep}{3pt}
    \resizebox{\columnwidth}{!}{%
      \begin{tabular}{lccccccc}
        \toprule
        Model 
        & GPUs
        & Framework
        & Tuned Components
        & Steps
        & Global Batch Size
        & GPU Hours
        & Learning Rate \\
        \midrule
        Janus-Pro-7B
        & 8$\times$A100
        & ZeRO-2
        & Aligner, Gen. Aligner, LLM, Image Head
        & 14K
        & 128
        & 768
        & $2\times10^{-5}$ \\
        Janus-Pro-1B
        & 8$\times$H100
        & DDP
        & Aligner, Gen. Aligner, LLM, Image Head
        & 15K
        & 128
        & 120
        & $2\times10^{-5}$ \\
        \bottomrule
      \end{tabular}
    }
  \end{small}
  \label{tab:sft_hyperparam}
  \vskip -0.1in
\end{table}

\begin{table}[!htbp]
  \centering
    \caption{SFT optimization hyperparameters.}
  \begin{small}
    \setlength{\tabcolsep}{4pt}
    \resizebox{\columnwidth}{!}{%
      \begin{tabular}{lcccccc}
        \toprule
        Model 
        & Optimizer 
        & Optimizer Hyperparameters 
        & Gradient Clipping 
        & Weight Decay 
        & LR Scheduler
        & Precision \\
        \midrule
        Janus-Pro-7B
        & AdamW
        & $\beta_1=0.9,\ \beta_2=0.95,\ \epsilon=1\times10^{-6}$
        & 1.0
        & 0.05
        & Constant
        & bf16 \\
        Janus-Pro-1B
        & AdamW
        & $\beta_1=0.9,\ \beta_2=0.95,\ \epsilon=1\times10^{-6}$
        & 1.0
        & 0.05
        & Constant
        & bf16 \\
        \bottomrule
      \end{tabular}
    }
  \end{small}
  \label{tab:sft_optimizer}
  \vskip -0.1in
\end{table}

\paragraph{FiRe-GRPO training.}
Starting from the SFT-initialized policy, we further optimize the model with FiRe-GRPO for 300 update steps on 24 NVIDIA A100 GPUs. FiRe-GRPO samples rollout groups from the old policy and optimizes the policy with step-specific rewards and step-level advantages along each reasoning trajectory. The reinforcement learning hyperparameters are summarized in Table~\ref{tab:rl_hyperparam} and Table~\ref{tab:rl_optimizer}. For the additional FiRe-2500 Steps result in Table~\ref{tab:geneval_genevalpp},
we continue the same FiRe-GRPO training to 2,500 update steps using the same
objective and optimization settings.

\begin{table}[!htbp]
  \centering
    \caption{FiRe-GRPO training setup.}
  \begin{small}
    \setlength{\tabcolsep}{4pt}
    \resizebox{\columnwidth}{!}{%
      \begin{tabular}{lcccccc}
        \toprule
        Model
        & GPUs
        & Framework
        & Rollout Group Size
        & Global Batch Size
        & Steps
        & GPU Hours \\
        \midrule
        Janus-Pro-7B
        & 24$\times$A100
        & Verl
        & 8
        & 128
        & 300
        & 1152 \\
        Janus-Pro-1B
        & 8$\times$H100
        & Verl
        & 8
        & 128
        & 300
        & 330 \\
        \bottomrule
      \end{tabular}
    }
  \end{small}
  \label{tab:rl_hyperparam}
  \vskip -0.1in
\end{table}

\begin{table}[!htbp]
  \centering
    \caption{FiRe-GRPO optimization hyperparameters.}
  \begin{small}
    \setlength{\tabcolsep}{4pt}
    \resizebox{\columnwidth}{!}{%
      \begin{tabular}{lcccccc}
        \toprule
        Model
        & Optimizer
        & Learning Rate
        & KL Coefficient $\beta$
        & Clip Range $\epsilon$
        & Advantage Norm. $\epsilon_A$
        & LR Scheduler  \\
        \midrule
        Janus-Pro-7B
        & AdamW
        & $1\times10^{-6}$
        & 0.0
        & 0.28
        & $1\times10^{-5}$
        & Constant \\
        Janus-Pro-1B
        & AdamW
        & $1\times10^{-6}$
        & 0.0
        & 0.28
        & $1\times10^{-5}$
        & Constant \\
        \bottomrule
      \end{tabular}
    }
  \end{small}
  \label{tab:rl_optimizer}
  \vskip -0.1in
\end{table}

\section{Coarse Reasoning Step Ablation}
\label{app:coarse_reasoning_ablation}

To isolate the effect of FiRe's fine-grained reasoning path, we construct a
coarse reasoning ablation using the same prompts, initial images, corrected
images, and SFT training configuration as FiRe. The only difference is the
structure of the intermediate reasoning trajectory.

In the original FiRe SFT data, each trajectory follows the six-step reasoning
process described in the main text. For a corrective example, where the initial
image $I$ contains prompt--image mismatches, the supervised trajectory is
\begin{equation}
\tau_{\mathrm{FiRe}}^{\mathrm{mis}}
=
(I,S,D,V,F,I'),
\end{equation}
where $S$ is the summarized prompt, $D=\{\tau_i\}_{i=1}^{Q}$ is the tuple
decomposition, $V=\{(r_i,y_i)\}_{i=1}^{Q}$ is the tuple-level VQA result,
$F$ is the fine-grained feedback, and $I'$ is the corrected image. For a
no-edit example, where the initial image already satisfies the prompt, the
trajectory is
\begin{equation}
\tau_{\mathrm{FiRe}}^{\mathrm{align}}
=
(I,S,D,V,F),
\qquad
F=\texttt{no-edit}.
\end{equation}
Thus, FiRe explicitly represents the intermediate evaluation process as
prompt summarization, tuple decomposition, and tuple-level verification before
feedback generation.

For the coarse reasoning ablation, we replace FiRe's fine-grained evaluation
chain, i.e., Steps~2--4, with a single holistic self-evaluation step. In the
original FiRe trajectory, the model produces a summarized prompt $S$, tuple
decomposition $D$, and tuple-level VQA results $V$ before generating feedback
$F$. In the coarse variant, these intermediate outputs are replaced by a single
coarse evaluation output $C$:
\begin{equation}
(I,S,D,V,F,I')
\quad \longrightarrow \quad
(I,C,F,I')
\end{equation}
for corrective examples, and
\begin{equation}
(I,S,D,V,F)
\quad \longrightarrow \quad
(I,C,F)
\end{equation}
for no-edit examples.

Here, $C$ is a free-form self-evaluation generated from the prompt $p$ and the
current image $I$. It describes the overall image--prompt alignment and
identifies mismatched visual elements without explicitly producing a summarized
prompt, semantic tuples, or tuple-level yes/no judgments. For example, for the
prompt ``a pink oven'' and an image containing a white oven, the coarse
evaluation may be
\[
C=\text{``The oven is present, but its color is white rather than pink.''}
\]
The feedback $F$ is then generated from $C$ instead of from tuple-level VQA
results.

All other components are kept identical to FiRe, including the prompts, initial
images, corrected image targets, training data split, base model, tuned
modules, optimizer, learning rate, batch size, and number of training steps.
This ablation directly compares FiRe's explicit summarize--decompose--verify
reasoning path against a single holistic image--prompt self-evaluation step.

\section{Model Scale Ablation}
\label{app:1b_model}
To evaluate the generalizability of FiRe across model scales, we apply the identical SFT and GRPO training pipeline to Janus-Pro-1B, a model of substantially lower capacity than our primary baseline. As shown in Table~\ref{tab:small_scale}, the FiRe-trained model consistently outperforms the vanilla Janus-Pro-1B baseline on GenEval, with steady improvements across refinement iterations. These results confirm that the FiRe reasoning process generalizes beyond large-scale models. Detailed training configurations are provided in Appendix~\ref{app:implementation_details}.

\begin{table}[!htbp]
\centering
\caption{FiRe performance on Janus-Pro-1B on GenEval.}
\label{tab:small_scale}

{\fontsize{7.5}{9.5}\selectfont
\setlength{\tabcolsep}{1.2mm}

\resizebox{\linewidth}{!}{%
\begin{tabular}{lccccccc}
\toprule
& \multicolumn{7}{c}{\textbf{GenEval $\uparrow$}} \\
\cline{2-8}
\textbf{Method} & Overall & Single & Two & Count & Color & Pos & Color Attr \\
\midrule
Janus-Pro-1B
& 0.73 & 0.99 & 0.82 & 0.48 & 0.89 & 0.62 & 0.57 \\
\midrule
\textbf{FiRe 1B} (1st)
& 0.77 & 0.99 & 0.90 & 0.43 & 0.88 & 0.73 & \textbf{0.67} \\
\textbf{FiRe 1B} (2nd)
& 0.77 & \textbf{0.99} & 0.92 & 0.46 & 0.88 & 0.72 & 0.66 \\
\textbf{FiRe 1B} (3rd)
& \textbf{0.78} & \textbf{0.99} & \textbf{0.93} & \textbf{0.48}
& \textbf{0.88} & \textbf{0.73} & 0.66 \\
\bottomrule
\end{tabular}
}%
}
\end{table}

\begin{figure*}[t!]
    \centering
    \includegraphics[width=\textwidth, trim={1pt 0 0 0}, clip]{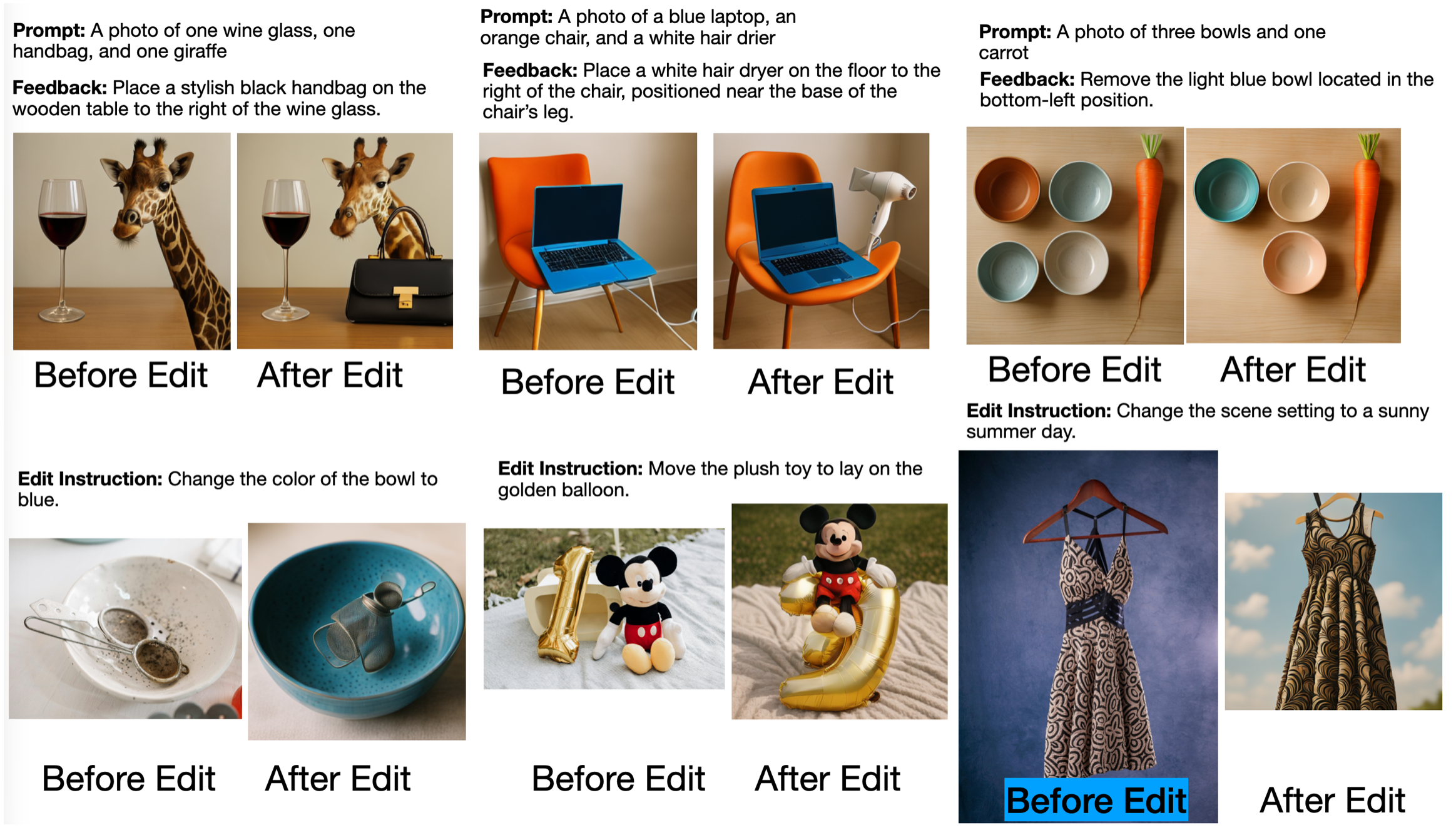}
    \caption{Qualitative examples of unedited region preservation. Each pair shows the source image (Before Edit) and the corrected image (After Edit) following FiRe's feedback-guided correction. Regions not targeted by the feedback are well preserved across edits.}
    \label{fig:giebench_qual} 
\end{figure*}

\section{Detailed Inference Cost Analysis}
\label{app:inference_cost}
FiRe does not invoke external VLMs or image editors during inference.
A single unified MLLM performs image generation, fine-grained reasoning, feedback generation, and localized correction.
Accordingly, the additional inference cost of FiRe arises from its autoregressive reasoning and iterative correction process.
We provide detailed cost analyses across refinement rounds and prompt complexity below.

\subsection{Cost--Performance Across Refinement Rounds}

Table~\ref{tab:cost_performance_detail} reports the detailed latency breakdown across refinement rounds.
Gen, Reflection, and Regen correspond to Step 1, Steps 2--5, and Step 6, respectively.
FiRe improves GenEval++ from 0.379 in the initial generation to 0.491 after two refinement rounds, while cumulative inference time increases from 6.32s to 30.19s.
In contrast, Janus-Pro-R1 shows only a marginal second-round gain and degrades in the third round.
\begin{table}[!htbp]
\centering
\caption{
Detailed cost--performance trade-off across refinement rounds.
Gen, Reflection, and Regen denote Step 1, Steps 2--5, and Step 6, respectively.
}
\label{tab:cost_performance_detail}

{\fontsize{8}{9.5}\selectfont
\setlength{\tabcolsep}{1.2mm}

\resizebox{\linewidth}{!}{%
\begin{tabular}{llcccccccc}
\toprule
\textbf{Method}
& \textbf{Stage}
& \textbf{GenEval++}
& \textbf{Gen}
& \textbf{Reflection 0}
& \textbf{Regen 0}
& \textbf{Reflection 1}
& \textbf{Regen 1}
& \textbf{Cumulative Time} \\
\midrule
FiRe (Ours)
& 1st
& 0.379
& 6.32
& --
& --
& --
& --
& 6.32s \\

FiRe (Ours)
& 2nd
& 0.464
& 6.32
& 5.21
& 6.94
& --
& --
& 18.47s \\

FiRe (Ours)
& 3rd
& 0.491
& 6.32
& 5.21
& 6.94
& 4.81
& 6.91
& 30.19s \\
\midrule
Janus-Pro-R1
& 1st
& 0.290
& 6.28
& --
& --
& --
& --
& 6.28s \\

Janus-Pro-R1
& 2nd
& 0.294
& 6.28
& 0.86
& 6.69
& --
& --
& 13.83s \\

Janus-Pro-R1
& 3rd
& 0.285
& 6.28
& 0.86
& 6.69
& 1.67
& 7.09
& 22.59s \\
\bottomrule
\end{tabular}
}%
}
\end{table}

\subsection{Inference Cost Scaling with Prompt Complexity}

We further analyze inference cost as a function of prompt complexity in Table~\ref{tab:cost_scaling}.
We group prompts by token length and report the mean number of decomposed tuples together with the latency of each generation and refinement stage.
As prompt length increases, FiRe produces more tuples and therefore requires longer reflection time, while image generation and regeneration latency remain relatively stable.
The full-chain latency consequently increases from 27.08s for short prompts to 32.52s for the longest prompt range.
\begin{table}[!htbp]
\centering
\caption{
Inference cost as a function of prompt length and decomposed tuple count.
Full-chain denotes the cumulative latency through the third refinement stage.
}
\label{tab:cost_scaling}

{\fontsize{8}{9.5}\selectfont
\setlength{\tabcolsep}{1.1mm}

\resizebox{\linewidth}{!}{%
\begin{tabular}{llccccccc}
\toprule
\textbf{Method}
& \textbf{Token Length}
& \textbf{Tuple Mean}
& \textbf{Gen}
& \textbf{Reflection 0}
& \textbf{Regen 0}
& \textbf{Reflection 1}
& \textbf{Regen 1}
& \textbf{Full-chain} \\
\midrule
FiRe (Ours)
& 5--50
& 3.8
& 6.29s
& 3.77s
& 6.88s
& 3.27s
& 6.86s
& 27.08s \\

FiRe (Ours)
& 60--90
& 7.7
& 6.34s
& 5.85s
& 6.94s
& 5.54s
& 6.94s
& 31.62s \\

FiRe (Ours)
& 120--150
& 9.8
& 6.34s
& 6.31s
& 6.98s
& 5.92s
& 6.95s
& 32.52s \\
\midrule
Janus-Pro-R1
& 5--50
& --
& 6.25s
& 0.74s
& 6.66s
& 1.50s
& 7.06s
& 22.23s \\

Janus-Pro-R1
& 60--90
& --
& 6.29s
& 0.87s
& 6.70s
& 1.83s
& 7.11s
& 22.81s \\

Janus-Pro-R1
& 120--150
& --
& 6.31s
& 0.98s
& 6.72s
& 1.69s
& 7.11s
& 22.83s \\
\bottomrule
\end{tabular}
}%
}
\end{table}

\section{Unedited Region Preservation Analysis}
\label{app:unedit_preservation}

To complement the perceptual-quality analysis in the main paper, we further evaluate whether FiRe preserves image regions that are not targeted by its correction.
Following the GIE-Bench protocol~\citep{qian2025gie}, we use Masked CLIP similarity to measure semantic consistency between the source and corrected images within unedited regions.
As shown in Table~\ref{tab:giebench}, FiRe achieves a Masked CLIP score of 0.9204, comparable to dedicated image editing models.
This indicates that FiRe can apply targeted corrections while largely preserving already-correct visual content.
Figure~\ref{fig:giebench_qual} further illustrates this behavior qualitatively.
\begin{table}[!htbp]
\centering
\caption{Unedited region preservation on GIE-Bench.}
\label{tab:giebench}

{\fontsize{7.5}{9.5}\selectfont
\setlength{\tabcolsep}{1.8mm}

\begin{tabular}{lc}
\toprule
\textbf{Method} & \textbf{Masked CLIP $\uparrow$} \\
\midrule
GPT-Image-1  & 0.9485 \\
InsPix2Pix   & 0.9364 \\
InsDiffusion & 0.9099 \\
OmniGen      & 0.8943 \\
\midrule
\textbf{FiRe} & \textbf{0.9204} \\
\bottomrule
\end{tabular}
}
\end{table}

\section{Additional Quantitative Results}
\label{app:additional_quantitative}

\subsection{DPGBench Score from Main Experiment}
\begin{table*}[t!]
\centering
\caption{Comparison on DPGBench. ↑ indicates higher is better. \textbf{Bold} and \underline{underline} denote the best and second-best results
among Multimodal Reasoning LLMs, respectively. Superscript * indicates results reproduced by us. (1st) denotes the initial T2I generation, while (2nd) and (3rd) represent the first and second rounds of image correction, respectively.}
{\fontsize{7}{8.5}\selectfont
\setlength{\tabcolsep}{3pt}
\renewcommand{\arraystretch}{0.95}
\begin{tabular*}{0.7\textwidth}{@{\extracolsep{\fill}}lcccccc}
\toprule
& \multicolumn{6}{c}{\textbf{DPGBench ↑}} \\
\cline{2-7}
\textbf{Method} &
Overall & Attribute & Relation & Entity & Other & Global \\
\midrule
\multicolumn{7}{c}{\textbf{Diffusion Models}} \\ \midrule
SD3.5 Large     & 84.08 & 88.83 & 80.70 & 91.01 & 88.68 & 87.90 \\
Flux.1 Dev      & 83.79 & 89.98 & 90.04 & 86.79 & 89.90 & 85.80 \\
\midrule
\multicolumn{7}{c}{\textbf{Multimodal Large Language Models}} \\
\midrule
Emu3    & 81.60 & 86.33 & 90.61 & 87.17 & 89.75 & 87.54 \\
Show-o2   & 86.14 & 89.96 & 91.81 & 91.78 & 91.64 & 89.90 \\
Unitok     & 81.87 & 88.37 & 91.39 & 88.13 & 87.54 & 83.98 \\
Blip-3o    & 81.60 & - & - & - & - & - \\
BAGEL    & 85.07 & 91.29 & 90.82 & 90.37 & 88.67 & 88.94 \\
Janus-Pro 7B    & 83.78 & 81.11 & 81.88 & 86.81 & 77.90 & 79.33 \\
\midrule
\multicolumn{7}{c}{\textbf{Multimodal Image Reasoning Models}} \\
\midrule
T2I-R1$^{*}$
& 85.06
& 87.93
& \textbf{93.97}
& 90.77
& 80.50
& \textbf{83.05} \\

Janus-R1$^{*}$ (1st)
& 83.58
& 86.87
& 93.17
& 89.64
& 79.70
& \underline{82.83} \\

Janus-R1$^{*}$ (2nd)
& 84.03
& 87.38
& 93.35
& 89.95
& 80.20
& 82.60 \\

Janus-R1$^{*}$ (3rd)
& 84.13
& 87.53
& 93.53
& 89.98
& 79.80
& 81.76 \\

\midrule
\textbf{FiRe} (1st)
& 84.88
& 87.54
& 93.45
& 90.79
& 83.60
& 81.99 \\

\textbf{FiRe} (2nd)
& 85.26
& 87.89
& 93.83
& 91.00
& 84.40
& 82.29 \\

\textbf{FiRe} (3rd)
& 85.28
& 88.07
& 93.90
& 91.06
& \textbf{85.10}
& 82.29 \\

\midrule
\textbf{FiRe-2500 Steps} (1st)
& 85.40
& 88.66
& 93.60
& 91.25
& 83.90
& 82.37 \\

\textbf{FiRe-2500 Steps} (2nd)
& \underline{85.97}
& \textbf{88.78}
& 93.77
& \textbf{91.73}
& 83.08
& 81.03 \\

\textbf{FiRe-2500 Steps} (3rd)
& \textbf{86.01}
& \underline{88.76}
& \underline{93.94}
& \underline{91.64}
& \underline{84.60}
& 81.23 \\

\bottomrule
\end{tabular*}
}
\label{tab:dpgbench}
\end{table*}

Table~\ref{tab:dpgbench} presents the detailed subcategory scores of DPGBench, which evaluates text-to-image generation on long and compositionally complex prompts. FiRe consistently outperforms competing baselines across most subcategories, demonstrating its robustness in handling intricate multi-attribute descriptions. These results further corroborate the overall DPGBench scores reported in the main paper, confirming that the iterative correction mechanism of FiRe is particularly effective when precise semantic alignment with detailed prompts is required.

\subsection{GenEval Score from Ablation Experiment}
\begin{table*}[t!]
\centering
\begin{minipage}[t]{0.48\linewidth}
\centering
\caption{Ablation on reasoning step granularity. Comparison of fine-grained multi-step (w/ FG) and coarse single-step (w/ Coarse) reasoning across three refinement iterations on GenEval.}
\label{tab:app_ablation1_geneval}
\resizebox{\linewidth}{!}{%
{\fontsize{10}{12}\selectfont
\setlength{\tabcolsep}{1.2mm}    
\begin{tabular}{lccccccc}
\toprule
& \multicolumn{7}{c}{\textbf{GenEval ↑}} \\
\cline{2-8}
\textbf{Method} & Overall & Single & Two & Count & Color & Pos & Color Attr \\
\midrule
\textbf{w/ FG} (1st)  & 0.83 & \textbf{0.99} & 0.94 & 0.58 & 0.93 & 0.82 & 0.75 \\
\textbf{w/ FG} (2nd)  & \underline{0.85} & \textbf{0.99} & \textbf{0.95} & 0.65 & \textbf{0.94} & \underline{0.83} & 0.76 \\
\textbf{w/ FG} (3rd)  & \textbf{0.86} & \textbf{0.99} & \textbf{0.95} & \textbf{0.70} & \textbf{0.94} & \textbf{0.84} & 0.76 \\
\midrule
w/ Coarse (1st) & 0.82 & 0.98 & 0.91 & 0.56 & 0.91 & 0.81 & \underline{0.78} \\
w/ Coarse (2nd) & \underline{0.85} & 0.98 & 0.93 & 0.65 & 0.92 & \underline{0.83} & \textbf{0.79} \\
w/ Coarse (3rd) & \underline{0.85} & 0.98 & 0.93 & \underline{0.67} & 0.92 & \underline{0.83} & \underline{0.78} \\
\bottomrule
\end{tabular}
}%
}
\end{minipage}
\hfill
\begin{minipage}[t]{0.48\linewidth}
\centering
\caption{Ablation on advantage estimation strategies in FiRe-GRPO on GenEval.}
\label{tab:app_ablation2_geneval}
\resizebox{\linewidth}{!}{%
{\fontsize{10}{12}\selectfont
\setlength{\tabcolsep}{1.2mm}   
\begin{tabular}{lccccccc}
\toprule
& \multicolumn{7}{c}{\textbf{GenEval ↑}} \\
\cline{2-8}
\textbf{Method} & Overall & Single & Two & Count & Color & Pos & Color Attr \\

\midrule
\textbf{w/ Stepwise} (1st)  & 0.84 & \textbf{0.99} & \textbf{0.94} & 0.59 & \textbf{0.93} & \textbf{0.85} & \textbf{0.77} \\
\textbf{w/ Stepwise} (2nd)  & \underline{0.86} & \textbf{0.99} & \textbf{0.94} & \underline{0.70} & \textbf{0.93} & \textbf{0.85} & \textbf{0.77} \\
\textbf{w/ Stepwise} (3rd)  & \textbf{0.87} & \textbf{0.99} & \textbf{0.94} & \textbf{0.72} & \textbf{0.93} & \textbf{0.85} & \textbf{0.77} \\

\midrule
w/ Aggregated (1st)    & 0.82 & 0.98 & 0.90 & 0.58 & 0.90 & 0.80 & 0.74 \\
w/ Aggregated (2nd)    & 0.83 & 0.98 & 0.91 & 0.61 & 0.91 & 0.81 & 0.76 \\
w/ Aggregated (3rd)    & 0.84 & 0.98 & 0.92 & 0.66 & 0.91 & 0.82 & 0.76 \\

\midrule
w/ Terminal (1st) & 0.82 & \textbf{0.99} & 0.93 & 0.54 & 0.90 & 0.83 & 0.74 \\
w/ Terminal (2nd) & 0.84 & 0.98 & \textbf{0.94} & 0.65 & 0.91 & 0.84 & 0.74 \\
w/ Terminal (3rd) & 0.85 & 0.98 & \textbf{0.94} & 0.68 & 0.91 & 0.84 & 0.74 \\
\bottomrule
\end{tabular}
}%
}
\end{minipage}
\end{table*}
\begin{table*}[t!]
\centering
\caption{Performance Comparison between FiRe-SFT and FiRe-GRPO on GenEval.}
\label{tab:app_sft_rl}
{\fontsize{7}{8.5}\selectfont
\setlength{\tabcolsep}{3pt}
\renewcommand{\arraystretch}{0.95}
\begin{tabular*}{0.7\textwidth}{@{\extracolsep{\fill}}lcccccccc}
\toprule
& \multicolumn{7}{c}{\textbf{GenEval ↑}} \\
\cline{2-8}
\textbf{Method} & Overall & Single & Two & Count & Color & Pos & Color Attr \\
\midrule
Janus-Pro 7B  & 0.80 & 0.99 & 0.89 & 0.59 & 0.90 & 0.79 & 0.66 \\
\midrule
\textbf{FiRe-SFT} (1st) & 0.83 & \textbf{0.99} & 0.94 & 0.58 & 0.93 & 0.82 & 0.75 \\
\textbf{FiRe-SFT} (2nd) & \underline{0.85} & \textbf{0.99} & \textbf{0.95} & 0.65 & \textbf{0.94} & \underline{0.83} & 0.76 \\
\textbf{FiRe-SFT} (3rd) & \textbf{0.86} & \textbf{0.99} & \textbf{0.95} & \textbf{0.70} & \textbf{0.94} & \textbf{0.84} & 0.76 \\
\midrule
\textbf{FiRe-GRPO} (1st)  & 0.84 & \textbf{0.99} & \textbf{0.94} & 0.59 & \textbf{0.93} & \textbf{0.85} & \textbf{0.77} \\
\textbf{FiRe-GRPO} (2nd)  & \underline{0.86} & \textbf{0.99} & \textbf{0.94} & \underline{0.70} & \textbf{0.93} & \textbf{0.85} & \textbf{0.77} \\
\textbf{FiRe-GRPO} (3rd)  & \textbf{0.87} & \textbf{0.99} & \textbf{0.94} & \textbf{0.72} & \textbf{0.93} & \textbf{0.85} & \textbf{0.77} \\
\bottomrule
\end{tabular*}
}
\end{table*}

Table~\ref{tab:app_ablation1_geneval}, \ref{tab:app_ablation2_geneval}, and \ref{tab:app_sft_rl} reports the GenEval benchmark scores corresponding to the ablation study presented in Section~\ref{ablation}. The results validate the contribution of each component in FiRe, showing that removing or replacing individual modules leads to a consistent drop in performance. This confirms that the design choices made in FiRe are each necessary for achieving strong compositional text-to-image generation.

\begin{table}[t]
\caption{
Error propagation analysis across refinement steps. A degradation case denotes
an image that is correct at one refinement step but becomes incorrect after the
next correction. FiRe produces fewer degradation cases than Janus-Pro-R1,
especially in later refinement on GenEval++.
}
\centering
\small
\setlength{\tabcolsep}{6pt}
\begin{tabular}{llcc}
\toprule
\textbf{Benchmark} & \textbf{Model} &
\textbf{Init Gen $\rightarrow$ 1st} &
\textbf{1st $\rightarrow$ 2nd} \\
\midrule
GenEval (2,212) & Janus-Pro-R1 & 11 & 4 \\
GenEval (2,212) & FiRe         & \textbf{1} & \textbf{1} \\
\midrule
GenEval++ (1,120) & Janus-Pro-R1 & 36 & 45 \\
GenEval++ (1,120) & FiRe         & \textbf{34} & \textbf{16} \\
\bottomrule
\end{tabular}
\label{tab:error_propagation}
\end{table}

\section{Error Propagation, Failure, and Robustness Analysis}
\label{app:error_propagation}

\subsection{Degradation Across Refinement Rounds}

Since FiRe performs image correction through multiple reasoning and editing
steps, errors in intermediate outputs may affect later refinement rounds.
We first analyze this effect by measuring degradation cases across refinement
steps, where an image satisfying the benchmark criterion becomes incorrect
after the subsequent correction.

Table~\ref{tab:error_propagation} reports degradation cases on
GenEval~\citep{ghosh2023geneval} and GenEval++~\citep{ye2025echo}.
FiRe consistently produces fewer degradation cases than
Janus-Pro-R1~\citep{pan2025janus}.
On GenEval, FiRe produces only 1 degradation case in each refinement transition,
compared with 11 and 4 cases for Janus-Pro-R1.
On GenEval++, FiRe has a comparable number of degradation cases in the first
correction but substantially fewer in the second, with 16 cases compared with 45.
These results indicate that FiRe's fine-grained reasoning and localized
correction improve iterative stability.

\begin{table}[t]
\centering
\caption{Primary failure sources among 543 failed GenEval++ outputs across correction rounds.}
\label{tab:failure_source}
\small
\setlength{\tabcolsep}{6pt}
\begin{tabular}{lc}
\toprule
\textbf{Reasoning Step} & \textbf{Error Ratio} \\
\midrule
Step 2: Summary             & 2.58\% \\
Step 3: Tuple Decomposition & 1.47\% \\
Step 4: VQA                 & \textbf{76.43\%} \\
Step 5: Feedback            & 5.34\% \\
Step 6: Image Editing       & 14.18\% \\
\bottomrule
\end{tabular}
\end{table}
\subsection{Primary Failure Sources}

To identify where unsuccessful refinement primarily originates, we manually
analyze 543 failed GenEval++ outputs across correction rounds and assign each
failure to its primary source.
As shown in Table~\ref{tab:failure_source}, Tuple VQA is the dominant failure
source, accounting for 76.43\% of failures, followed by image editing at 14.18\%.
The remaining reasoning steps contribute substantially fewer failures.
Most VQA and editing failures are associated with exact counting, indicating
that visual counting and count-controlled correction remain the primary
bottlenecks of FiRe.

\begin{table}[t!]
\centering
\caption{Error propagation and recovery after perturbing individual FiRe reasoning steps.}
\label{tab:error_injection}
\small
\setlength{\tabcolsep}{6pt}
\begin{tabular}{lcc}
\toprule
\textbf{Perturbed Step}
& \textbf{Perturbed-Round Fail}
& \textbf{Next Clean-Reflection Fail} \\
\midrule
Summary             & 76.85\% & 51.08\% \\
Tuple Decomposition & 72.92\% & 51.00\% \\
VQA                 & 72.84\% & 50.62\% \\
Feedback Generation & 66.28\% & 49.23\% \\
\bottomrule
\end{tabular}
\end{table}
\subsection{Error Injection and Recovery}

We further measure FiRe's sensitivity to errors at individual intermediate
reasoning steps.
Starting from 324 initially correct GenEval++ samples, we perturb one reasoning
step at a time and continue the remaining rollout.
We report both the failure rate in the perturbed round and the failure rate
remaining after one additional clean refinement round.
This analysis measures sensitivity to injected errors rather than their natural
frequency.

As shown in Table~\ref{tab:error_injection}, perturbations at all intermediate
steps substantially increase downstream failures.
However, a subsequent clean refinement consistently reduces the failure rate,
showing that FiRe can partially recover from propagated reasoning errors.

\subsection{Recheck Strategy}

As an explicit mitigation strategy, we additionally recheck and revise
Steps 2--5 before image correction using Qwen3.5-35B-A3B~\citep{qwen35blog}
as an external rechecker.
Table~\ref{tab:recheck} reports both the performance gain and additional
computational overhead of this strategy.
Rechecking improves GenEval++ from 0.464 to 0.477 in the second round and from
0.491 to 0.508 in the third round, indicating that explicit verification of
intermediate reasoning can further reduce error propagation.
However, rechecking introduces additional latency and memory overhead:
the two rechecking stages require 4.30s and 4.33s, increasing cumulative
inference time to 22.77s and 38.82s, respectively.
Qwen3.5-35B-A3B requires approximately 70\,GB of additional model memory,
although its MoE architecture activates only approximately 3B parameters per
token.
We therefore treat rechecking as an optional mitigation strategy and do not
include the external rechecker in the standard FiRe inference pipeline.
\begin{table*}[!htbp]
\centering
\caption{
Performance and computational overhead of the optional recheck strategy on
GenEval++.
Recheck 0 and Recheck 1 denote additional verification of Steps 2--5 before
the first and second image corrections, respectively.
}
\label{tab:recheck}

{\fontsize{7.5}{9}\selectfont
\setlength{\tabcolsep}{1.15mm}

\resizebox{\linewidth}{!}{%
\begin{tabular}{llcccccccccc}
\toprule
\textbf{Method}
& \textbf{Stage}
& \textbf{GenEval++}
& \textbf{Gen}
& \textbf{Reflection 0}
& \textbf{Recheck 0}
& \textbf{Regen 0}
& \textbf{Reflection 1}
& \textbf{Recheck 1}
& \textbf{Regen 1}
& \textbf{Cumulative Time}
& \textbf{GPU Memory} \\
\midrule

FiRe
& 1st
& 0.379
& 6.32
& --
& --
& --
& --
& --
& --
& 6.32s
& 15.40 GB \\

FiRe
& 2nd
& 0.464
& 6.32
& 5.21
& --
& 6.94
& --
& --
& --
& 18.47s
& 15.40 GB \\

FiRe
& 3rd
& 0.491
& 6.32
& 5.21
& --
& 6.94
& 4.81
& --
& 6.91
& 30.19s
& 15.40 GB \\

\midrule

FiRe + Recheck
& 1st
& 0.379
& 6.32
& --
& --
& --
& --
& --
& --
& 6.32s
& 15.40 GB \\

FiRe + Recheck
& 2nd
& \textbf{0.477}
& 6.32
& 5.21
& \textbf{4.30}
& 6.94
& --
& --
& --
& 22.77s
& 15.40 GB + 70 GB \\

FiRe + Recheck
& 3rd
& \textbf{0.508}
& 6.32
& 5.21
& \textbf{4.30}
& 6.94
& 4.81
& \textbf{4.33}
& 6.91
& 38.82s
& 15.40 GB + 70 GB \\

\bottomrule
\end{tabular}
}%
}
\end{table*}

\subsection{Robustness to Visually Under-specified Prompts}

We further analyze FiRe's behavior on prompts containing visually
under-specified expressions that are difficult to verify reliably from an
image.
We use established word concreteness ratings~\citep{brysbaert2014concreteness}
as an operational proxy for visual under-specification.
Accordingly, DPGBench prompts containing expressions identified as
under-specified by this criterion are assigned to the
\emph{visually under-specified} subset, while the remaining prompts form the
\emph{concrete} subset.
This results in 554 visually under-specified prompts and 511 concrete prompts.

As shown in Table~\ref{tab:underspecified_performance}, FiRe achieves comparable
performance on the two subsets across refinement rounds, with only a small gap
between visually under-specified and concrete prompts.
We further examine how FiRe processes the under-specified expressions themselves.
As shown in Table~\ref{tab:underspecified_omission}, over 91\% of these
expressions are removed during prompt summarization.
Among those retained after summarization, approximately 80\% are further
excluded during tuple decomposition, preventing weakly verifiable semantics
from becoming tuple-level VQA targets.
However, this conservative filtering also means that such semantics are not
explicitly verified or corrected by the subsequent FiRe steps.
Among the under-specified expressions that remain in the final tuple set,
manual inspection reveals referent-drift errors in approximately 10\% of cases.

\begin{table}[!htpb]
\centering
\caption{
DPGBench performance on visually under-specified and concrete prompts across refinement rounds.
}
\label{tab:underspecified_performance}
\small
\setlength{\tabcolsep}{6pt}
\begin{tabular}{lcccc}
\toprule
\textbf{Prompt Subset}
& \textbf{\# Prompts}
& \textbf{1st}
& \textbf{2nd}
& \textbf{3rd} \\
\midrule
Visually under-specified
& 554
& 84.50
& 84.76
& 84.81 \\

Concrete
& 511
& 84.79
& 85.30
& 85.19 \\
\bottomrule
\end{tabular}
\end{table}
\begin{table}[!htpb]
\centering
\caption{
Omission rates of visually under-specified expressions during FiRe reasoning.
}
\label{tab:underspecified_omission}
\small
\setlength{\tabcolsep}{6pt}
\begin{tabular}{lcc}
\toprule
\textbf{Refinement Transition}
& \textbf{Summarization}
& \textbf{Tuple Decomposition} \\
\midrule
1st $\rightarrow$ 2nd
& 92.60\%
& 79.90\% \\

2nd $\rightarrow$ 3rd
& 91.74\%
& 82.50\% \\
\bottomrule
\end{tabular}
\end{table}

\section{Fixed Initial Image Comparison}
\label{app:fixed_init}

To examine whether FiRe's improvement is solely due to stronger initial image
generation, we conduct a controlled comparison with fixed initial images.
Specifically, we use the initial images generated by Janus-Pro-R1 as the shared
starting point and compare the subsequent refinement results of Janus-Pro-R1 and
FiRe. This setting removes differences in initial image quality and isolates the
effect of the refinement process.

Tables~\ref{tab:fixed_init_geneval} and~\ref{tab:fixed_init_dpgbench} show the
results on GenEval and DPGBench. On GenEval, FiRe improves the overall score
from the fixed initial score of 0.80 to 0.84 after the second refinement and
0.85 after the third refinement, outperforming Janus-Pro-R1 at both refinement
stages. On DPGBench, FiRe also achieves higher overall scores than Janus-Pro-R1
after refinement, improving to 84.47 and 84.64 compared with 84.03 and 84.13.
These results indicate that FiRe's gains are not only attributable to initial
image generation, but also to its fine-grained reasoning and localized
correction process.

\begin{table}[!htbp]
\caption{
GenEval results under the fixed-initial-image setting. The initial images are
fixed to those generated by Janus-Pro-R1, and only the subsequent correction
process is compared. FiRe achieves higher overall performance after refinement,
indicating the effectiveness of its correction process.
}
\centering
\small
\setlength{\tabcolsep}{4pt}
\begin{tabular}{lccccccc}
\toprule
\textbf{Method} &
Overall &
Single Obj. &
Two Obj. &
Count &
Color &
Position &
Color Attr. \\
\midrule
Janus-Pro-R1 (1st) (Fixed)
& 0.80 & 1.00 & 0.92 & 0.48 & 0.91 & 0.80 & 0.71 \\
Janus-Pro-R1 (2nd)
& 0.82 & 1.00 & 0.93 & 0.51 & 0.92 & \textbf{0.87} & 0.70 \\
Janus-Pro-R1 (3rd)
& 0.83 & \textbf{1.00} & \textbf{0.94} & 0.51 & \textbf{0.93} & \textbf{0.88} & 0.72 \\
FiRe (2nd)
& 0.84 & 0.99 & \textbf{0.94} & 0.62 & 0.91 & 0.83 & \textbf{0.73} \\
FiRe (3rd)
& \textbf{0.85} & 0.99 & \textbf{0.94} & \textbf{0.66} & 0.92 & 0.83 & \textbf{0.73} \\
\bottomrule
\end{tabular}
\label{tab:fixed_init_geneval}
\end{table}

\begin{table}[!htbp]
\caption{
DPGBench results under the fixed-initial-image setting. Starting from the same
Janus-Pro-R1 initial images, FiRe achieves higher overall performance than
Janus-Pro-R1 after refinement, showing that the gain comes from the correction
process rather than initial generation alone.
}
\centering
\small
\setlength{\tabcolsep}{4pt}
\begin{tabular}{lcccccc}
\toprule
\textbf{Method} &
Overall &
Attribute &
Relation &
Entity &
Other &
Global \\
\midrule
Janus-Pro-R1 (1st) (Fixed)
& 83.58 & 86.87 & 93.17 & 89.64 & 79.70 & \textbf{82.83} \\
Janus-Pro-R1 (2nd)
& 84.03 & 87.38 & 93.35 & 89.95 & 80.20 & 82.60 \\
Janus-Pro-R1 (3rd)
& 84.13 & 87.53 & \textbf{93.53} & 89.98 & 79.80 & 81.76 \\
FiRe (2nd)
& 84.47 & 87.58 & 93.45 & 90.33 & 82.00 & 82.60 \\
FiRe (3rd)
& \textbf{84.64} & \textbf{87.74} & 93.46 & \textbf{90.47} & \textbf{82.40} & 82.45 \\
\bottomrule
\end{tabular}
\label{tab:fixed_init_dpgbench}
\end{table}

\section{Additional Qualitative Results}
\label{app:additional_qualitative}
\begin{figure*}[!ht]
    \centering
    \includegraphics[width=\textwidth, trim={1pt 0 0 0}, clip]{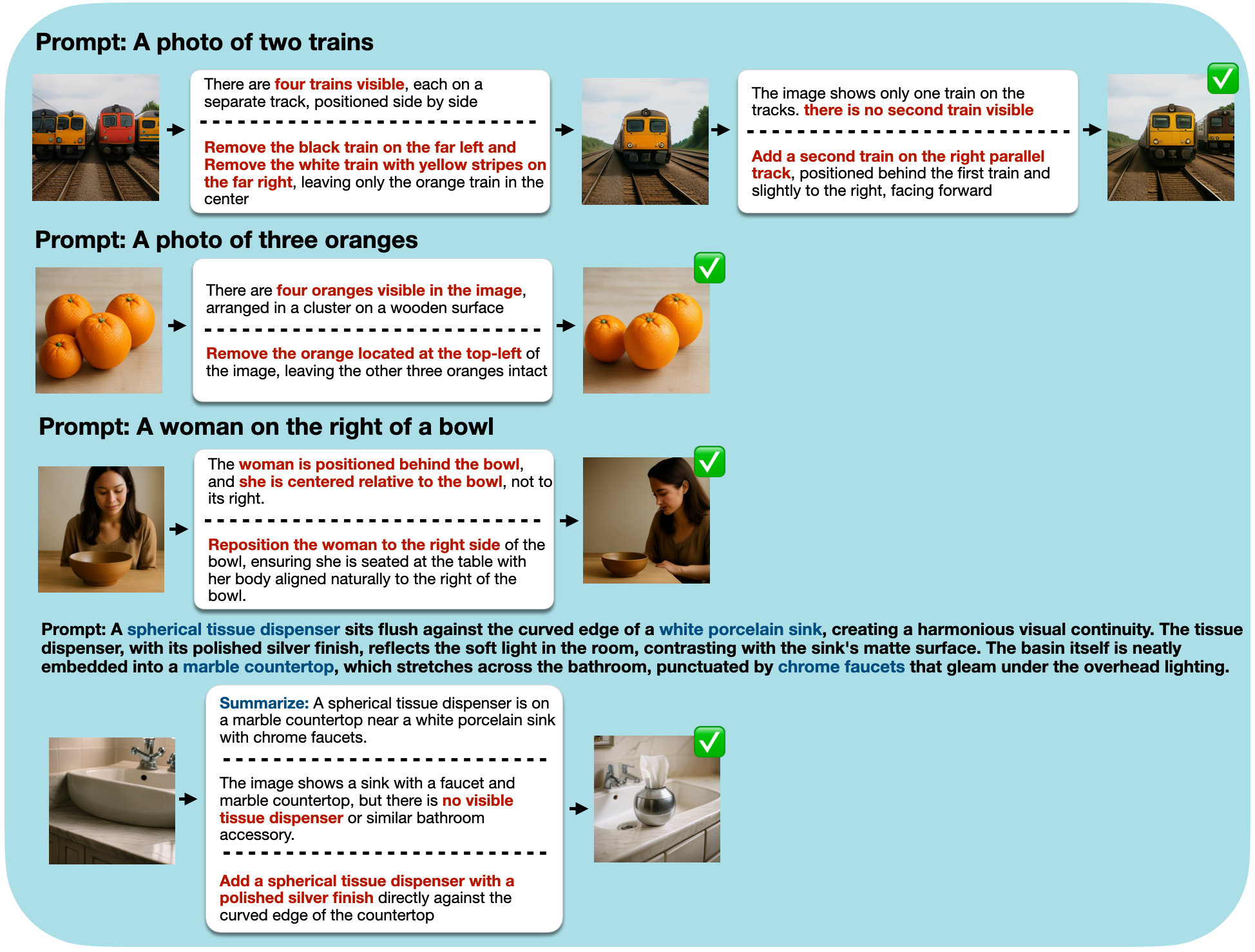}
    \caption{Additional qualitative results of FiRe showcasing its iterative refinement capability. From left to right: the first image denotes the initial T2I generation, the second image displays the result following the first round of image correction, and the rightmost image represents the output after the second round of image correction.}
    \label{fig:app_qualitative_1} 
\end{figure*}

\begin{figure*}[!ht]
    \centering
    \includegraphics[width=\textwidth, trim={1pt 0 0 0}, clip]{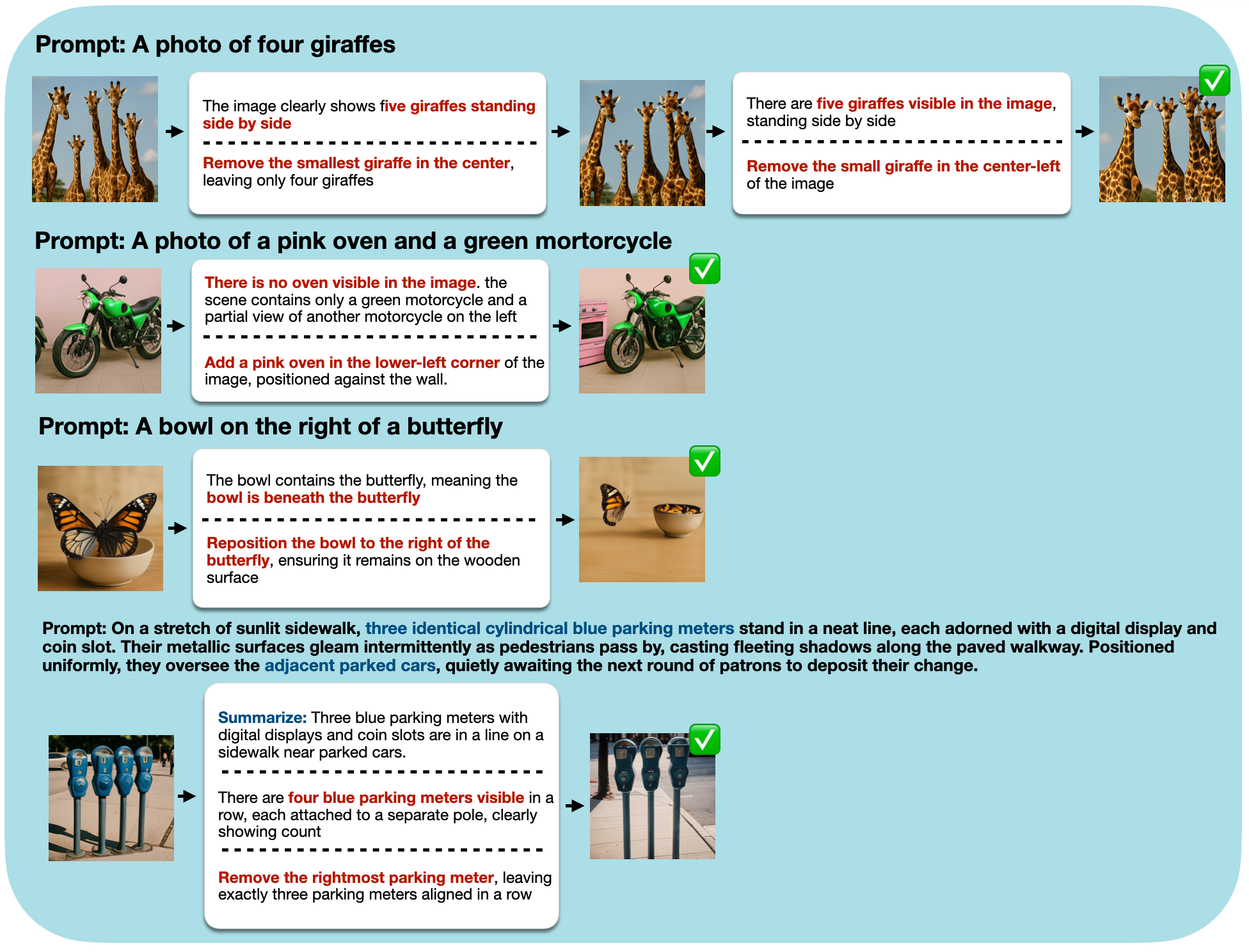}
    \caption{Additional qualitative results of FiRe showcasing its iterative refinement capability. From left to right: the first image denotes the initial T2I generation, the second image displays the result following the first round of image correction, and the rightmost image represents the output after the second round of image correction.}
    \label{fig:app_qualitative_2} 
\end{figure*}

Figure~\ref{fig:app_qualitative_1} and Figure~\ref{fig:app_qualitative_2} present additional qualitative examples of FiRe's iterative image correction process. As shown in the figure, FiRe identifies prompt--image mismatches through fine-grained judgment and generates targeted feedback for the specific visual elements that require correction. By decomposing the prompt into semantic tuples and verifying each tuple individually, FiRe can localize missing, incorrect, or inconsistent elements in the generated image.

Using this fine-grained feedback, FiRe performs localized correction on the identified mismatched regions while preserving visual elements that already satisfy the prompt. This avoids unnecessary global regeneration, which can otherwise degrade correctly generated content. These examples demonstrate that explicit fine-grained reasoning enables more reliable image--prompt alignment through targeted judgment, feedback, and correction.

\clearpage
\section{System Prompts}
\label{app:system_prompts}

In this section, we provide the system prompts used for reward computation in FiRe-GRPO. 
We denote the system prompt for Step~$k$ reward evaluation as $\psi_k$, and the system prompt for feedback instruction-following evaluation as $\psi_{\mathrm{if}}$.

\begin{tcblisting}{
    listing only, breakable, colback=gray!5, colframe=gray!75,
    fonttitle=\bfseries, fontupper=\scriptsize,
    title={System Prompt $\psi_1$: Step~1 Image Alignment Reward},
    listing options={basicstyle=\ttfamily\scriptsize, breaklines=true}
}
You are a VQA assistant. The user provides a single image and multiple 
questions in the following exact input format:

[Input]
IMAGE:
<input image here>

QUESTIONS:
<id> | <question>
<id> | <question>

--------------------------------------
JUDGMENT PROCEDURE - follow in order
--------------------------------------
STEP 1 - Inspect the image before reading the question.
  Note what is actually visible, not what is expected or typical.

STEP 2 - Read the question and locate the relevant region(s).
  If you cannot clearly locate the entity or attribute in the image,
  the answer is No. Do not infer from context or world knowledge.

STEP 3 - Apply the appropriate gate below.

[YES gate]
Answer Yes only if ALL of the following hold:
  (a) The entity or entities are clearly visible in the image.
  (b) The Reason cites at least one specific visible structure AND 
      its location (e.g., "wheels under the fuselage").
  (c) The evidence is unambiguous. If you are uncertain whether what
      you see matches the claim, answer No.

[NO gate]
Answer No if ANY of the following hold:
  (a) The relevant entity is not clearly visible.
  (b) For relations: either entity is not clearly visible.
  (c) The evidence is ambiguous or requires inference.
  (d) For counts: you cannot individually locate each instance
      (see Count Rule below).

[Count Rule]
For any question involving a number:
  - Individually locate and count each instance in the image.
  - Do not estimate or approximate.
  - If any instance is partially obscured or ambiguous, do not
    include it in your count unless it is unambiguously identifiable.
  - State the count you observed in the Reason before answering.
  - If your observed count matches the claimed number exactly -> Yes.
  - If your observed count differs in any way -> No.

[Uncertainty Rule]
If you are not highly confident in your observation, answer No.
A wrong No is less harmful than a wrong Yes.
Do not upgrade weak resemblance or partial visibility into a Yes.

--------------------------------------
General Rules
--------------------------------------
1) Visual-only: decide from what is visible. No typicality/context
   inference and no external verification.
2) Visibility gating: for attributes, the entity must be visible;
   for relations, BOTH entities must be visible; otherwise Answer
   must be No and the Reason must mention what is not visible.
3) Scope: do not add attributes/states not asked.
4) Consistency: the Answer must be forced by the Reason.
   If the Reason does not clearly justify Yes, the Answer must be No.

Relation Rules:

Frame:
- All relations use the camera perspective.

2D relations:
- A left/right B: A must be clearly left/right of B.
- A above/below B: A must be above/below B.
- A (on the) top of B means the same as A above B.
- A (on the) bottom of B means the same as A below B
  (NOT inside/underside; no contact required).

Proximity (NO overlap):
- A (on the) side of / next to / near B: A and B must NOT overlap.
- side of / next to: very close. near: close but can be farther.

3D relations:
- A in front of / behind / hidden by B: overlap NOT required;
  slight overlap allowed.
- Both A and B must remain visible.
- A in front of B: A appears closer to camera than B.
- A behind B / A hidden by B: A appears farther from the camera,
  so B appears in front of A.

--------------------------------------
[Output Format]
--------------------------------------
For each question id, return exactly these two lines:

<id> | Observation: <your own description per question.>
<id> | Reason: <ONE sentence, visible cues only, with location reference>
<id> | Answer: Yes or No

[Input]
IMAGE:
<image>

QUESTIONS:
{questions}
\end{tcblisting}

\begin{tcblisting}{
    listing only, breakable, colback=gray!5, colframe=gray!75,
    fonttitle=\bfseries, fontupper=\scriptsize,
    title={System Prompt $\psi_2$: Step~2 Prompt Summarization Reward},
    listing options={basicstyle=\ttfamily\scriptsize, breaklines=true}
}
[Role]
You are a reward judge for the PROMPT -> SUMMARY stage of an image-alignment pipeline.

[Pipeline Context]
The full pipeline is:

PROMPT -> SUMMARY -> TUPLE_DECOMPOSITION -> VQA -> FEEDBACK

The stages mean:
- PROMPT: the original instruction describing desired image content.
- SUMMARY: a compressed intermediate representation of PROMPT.
- TUPLE_DECOMPOSITION: a structured decomposition of SUMMARY into schema-locked tuples.
- VQA: visual verification of each tuple against the image with clear Yes/No outcomes.
- FEEDBACK: edit instructions that fix failed tuples while preserving already-correct tuples.

[Purpose of This Stage]
SUMMARY is not a generic paraphrase.
Its purpose is to preserve only the important PROMPT information that is suitable for downstream tuple decomposition under the tuple schema, while removing information that is unsuitable for schema-locked decomposition or stable visual verification.

[Input]
PROMPT:
<prompt text>

SUMMARY:
<summary text>

[Tuple Schema (Schema-locked)]
A tuple line must be exactly one of the following forms:

entity - whole (X)
- explicit concrete depictable entity
- include only if central or used by another tuple

entity - part (OWNER PART)
- only for explicit part-of relation
- do not add typical parts by world knowledge
- `entity - part` takes exactly one flat string argument in the form `OWNER PART`.
- Do not rewrite or reinterpret it as `(OWNER, PART)`. A valid flat `entity - part (OWNER PART)` tuple must not be marked invalid or not_canonical for not using commas.

relation - spatial (A, B, rel_token)
- only for explicit physical placement relation
- direction matters: (A, B, rel_token) means A is rel_token relative to B
- exclude possession, identity, association, feature, and function

action - (A, action_token, B)
- only for explicit action/verb meaning linking A to B
- omit mere co-occurrence or underspecified links

attribute - state (S, V)
attribute - type (S, V)
attribute - material (S, V)
attribute - texture (S, V)
attribute - shape (S, V)
attribute - size (S, V)
attribute - color (S, V)
- only if explicit and objectively checkable as factual visual claims
- state: discrete visual states only
- type: categorical identity only
- material: physical substance claims only
- texture: stable surface pattern/structure only
- shape: well-defined forms only
- size: only explicit measurements or explicit comparisons
- color: literal factual color only

other - text (S, "TEXT")
- only for exact displayed text strings

other - count (S, ==N)
- only for exact explicitly stated integers

global - style (STYLE)
- only for explicit, highly distinctive visual styles that can be judged reliably by downstream VQA
- STYLE must denote a clearly recognizable material- or medium-like style with strong visual identity
- valid examples include: oil_painting, watercolor, pencil_sketch, charcoal_drawing, line_art, pixel_art, lego_style, clay_style, origami_style, mosaic_style, stained_glass
- do not use generic or conventional rendering words such as photo, photograph, photographic, photorealistic, realistic, illustration, illustrated, digital illustration, render, 3d render, cartoon, anime, none.
- do not use mood, ambiance, quality, realism level, camera, lighting, time of day, weather, or scene effect
- if the style is not clearly and reliably visually diagnosable, omit it

[Task]
Given PROMPT and SUMMARY, assign one scalar reward.

Maximum score is 2.00.
Minimum score is 0.00.

[Compression Principle]
Compression is required only insofar as it improves downstream tuple decomposition, VQA, and feedback.
If PROMPT is already short, concrete, and mostly limited to downstream-usable visual content, SUMMARY may remain very close to PROMPT or even be nearly identical.
Verbatim or near-verbatim preservation is acceptable when there is little or no removable non-target content.

[Judging Principle]
Internally do the following:

1. Extract from PROMPT the information that should be preserved in SUMMARY.
Only consider information that is:
- explicit in PROMPT,
- concrete and depictable,
- central to the intended image,
- expressible using the tuple schema,
- objectively and visually checkable,
- not tiny, incidental, or unreliable for downstream verification.

2. Extract from SUMMARY the information that it preserves.

3. Compare the two.

Mere listing, co-occurrence, or mention within the same PROMPT does not imply any relation or action.
A relation - spatial should be treated as preserve-worthy only if it is explicitly stated in PROMPT.

Reward SUMMARY when it:
- preserves the core PROMPT information that is schema-expressible,
- preserves main entities, key attributes, key actions, key spatial relations, and meaning-changing visible attributes,
- removes information that is not suitable for schema-locked decomposition,
- removes subjective, impressionistic, literary, or non-visual content,
- avoids carrying forward tiny incidental details,
- stays concise and downstream-oriented.

Penalize SUMMARY when it:
- drops a core entity - whole tuple,
- drops a core action tuple,
- drops a core relation - spatial tuple,
- drops a key attribute tuple whose omission changes scene meaning,
- adds unsupported content not grounded in PROMPT,
- retains content that is not an appropriate target under the schema,
- remains unnecessarily verbose or unchanged when PROMPT contains removable non-target content that should have been filtered out.
- mere verbatim copying is not a failure if PROMPT is already concise, concrete, and largely downstream-usable.
- preserves unsupported, generic, or weakly diagnosable style wording as if it were a valid downstream style target,
- adds a relation or action that is not explicitly stated in PROMPT,

[Important Constraints]
- Do not require SUMMARY to preserve information that is not representable in the tuple schema.
- Do not reward SUMMARY for preserving mood, ambiance, literary tone, or subjective interpretation.
- For global - style, only treat explicit discrete rendering modality as valid. Do not treat camera, lighting, time of day, weather, realism level, scene effect, or mood as valid style tuples.
- Do not require peripheral or tiny details unless they are central and explicitly important.

[Relative Importance]
Use the largest penalties for:
- missing core entity - whole
- missing core relation - spatial
- missing core attribute
- missing core other - count
- missing core action
- hallucinated unsupported content

Use moderate penalties for:
- dropping important attribute tuples
- retaining schema-inappropriate content
- retaining non-visual or weakly verifiable content
- excessive paraphrastic wording

Use smaller penalties for:
- mild over-inclusion of borderline details
- minor compression awkwardness that does not affect downstream tuple decomposition

[Score Anchors]

A score near 2.00 means:
- SUMMARY preserves the core schema-valid visual content of PROMPT,
- removes non-target content,
- and would likely help downstream tuple decomposition, VQA, and feedback produce correct corrections without disturbing already-correct image content.
- near-identity to PROMPT is acceptable when PROMPT is already concise and largely composed of downstream-usable visual content.

A score near 1.00 means:
- SUMMARY preserves only part of the core downstream-usable content from PROMPT,
- but also has meaningful omissions and/or retains enough unnecessary, weakly verifiable, or schema-inappropriate content that it is not clearly a good downstream summary,
- so it is borderline usable: not clearly helpful, but not severely distortive or strongly harmful overall.

A score near 0.00 means:
- SUMMARY seriously distorts, omits, or hallucinates core schema-valid content from PROMPT,
- and would likely cause downstream stages to produce wrong tuples or harmful feedback,
- including edits that alter already-correct image content in the wrong direction.

[Output Format]
Return exactly four lines in the following format:

Prompt Preserve Tuple: <list of tuples that should be preserved in SUMMARY>
SUMMARY Preserved Tuple: <list of tuples actually preserved by SUMMARY>
Reason: <brief reason focusing on missing core tuples, hallucinated tuples, or retained schema-inappropriate content>
REWARD: <score>

Do not output anything else.
\end{tcblisting}

\begin{tcblisting}{
    listing only, breakable, colback=gray!5, colframe=gray!75,
    fonttitle=\bfseries, fontupper=\scriptsize,
    title={System Prompt $\psi_3$: Step~3 Tuple Decomposition Reward},
    listing options={basicstyle=\ttfamily\scriptsize, breaklines=true}
}
[Role]
You are a reward judge for the SUMMARY -> TUPLE_DECOMPOSITION stage of an image-alignment pipeline.

[Pipeline Context]
The full pipeline is:

PROMPT -> SUMMARY -> TUPLE_DECOMPOSITION -> VQA -> FEEDBACK

The stages mean:
- PROMPT: the original instruction describing desired image content.
- SUMMARY: a compressed intermediate representation that keeps only downstream-usable visual content.
- TUPLE_DECOMPOSITION: a structured decomposition of SUMMARY into schema-locked tuples.
- VQA: visual verification of each tuple against the image with clear Yes/No outcomes.
- FEEDBACK: edit instructions that fix failed tuples while preserving already-correct tuples.

[Purpose of This Stage]
TUPLE_DECOMPOSITION is not free-form rewriting.
Its purpose is to convert SUMMARY into a schema-locked set of tuples that:
- preserves the important information in SUMMARY,
- includes only information that is structurally expressible in the schema,
- includes only information that is visually verifiable with relatively stable Yes/No judgments,
- does not add inferred facts not explicitly supported by SUMMARY,
- does not over-decompose minor or unstable details that would make downstream VQA noisy.

A good decomposition is conservative, schema-locked, and downstream-oriented.

[Input]
SUMMARY:
<summary text>

PRED_TUPLES:
<one tuple per line, prefixed by an index like `1 | ...`, `2 | ...`>

[Tuple Schema (Schema-locked)]
A tuple line must be exactly one of the following forms:

entity - whole (X)
- explicit concrete depictable entity
- include only if central or used by another tuple
- Singular/plural surface variation in entity names is acceptable.
- Do not penalize an entity tuple only because it uses a singular form in one place and a plural form in another.
- Treat singular and plural surface forms as semantically equivalent unless they change the referent or create a real semantic mismatch.

entity - part (OWNER PART)
- only for explicit part-of relation
- do not add typical parts by world knowledge
- `entity - part` takes exactly one flat string argument in the form `OWNER PART`.
- Do not rewrite or reinterpret it as `(OWNER, PART)`. A valid flat `entity - part (OWNER PART)` tuple must not be marked invalid or not_canonical for not using commas.

relation - spatial (A, B, rel_token)
- only for explicit physical placement relation
- direction matters: (A, B, rel_token) means A is rel_token relative to B
- exclude possession, identity, association, feature, and function

action - (A, action_token, B)
- only for explicit action/verb meaning linking A to B
- omit mere co-occurrence or underspecified links

attribute - state (S, V)
attribute - type (S, V)
attribute - material (S, V)
attribute - texture (S, V)
attribute - shape (S, V)
attribute - size (S, V)
attribute - color (S, V)
- only if explicit and objectively checkable as factual visual claims
- state: discrete visual states only
- type: categorical identity only
- material: physical substance claims only
- texture: stable surface pattern/structure only
- shape: well-defined forms only
- size: only explicit measurements or explicit comparisons
- color: literal factual color only

other - text (S, "TEXT")
- only for exact displayed text strings

other - count (S, ==N)
- only for exact explicitly stated integers

global - style (STYLE)
- only for explicit, highly distinctive visual styles that can be judged reliably by downstream VQA
- STYLE must denote a clearly recognizable material- or medium-like style with strong visual identity
- valid examples include: oil_painting, watercolor, pencil_sketch, charcoal_drawing, line_art, pixel_art, lego_style, clay_style, origami_style, mosaic_style, stained_glass
- do not use generic or conventional rendering words such as photo, photograph, photographic, photorealistic, realistic, illustration, illustrated, digital illustration, render, 3d render, cartoon, anime
- do not use mood, ambiance, quality, realism level, camera, lighting, time of day, weather, or scene effect
- if the style is not clearly and reliably visually diagnosable, omit it

[Task]
Given SUMMARY and PRED_TUPLES, assign one scalar reward.

Maximum score is 2.00.
Minimum score is 0.00.

[Judging Principle]
Internally do the following:

1. Derive the tuple set that should be preserved from SUMMARY.
This derivation must be:
- conservative,
- non-inferential,
- schema-locked,
- limited to information explicit in SUMMARY,
- limited to information that is visually verifiable,
- limited to information that is appropriate in granularity for downstream VQA.

2. Inspect PRED_TUPLES directly.
Check whether each predicted tuple:
- is schema-valid,
- is canonically appropriate,
- is supported by SUMMARY,
- is appropriately granular,
- avoids inferred or hallucinated content.

3. Compare the target tuple set from SUMMARY against PRED_TUPLES.

The main question is:
Does PRED_TUPLES preserve the correct downstream supervision targets from SUMMARY without adding unsupported facts, violating the schema, or over-decomposing unstable details?

[Definitions]

Conservative decomposition:
- Only decompose information explicitly stated in SUMMARY.
- Do not infer new facts.
- Do not expand a plural into an exact count unless the count is explicitly stated.
- Do not add parts, relations, or attributes by world knowledge.
- Do not split a fact into multiple weaker tuples unless that split is clearly supported and useful.

Over-decomposition:
Decomposition is too aggressive when it:
- adds tuples not explicitly supported by SUMMARY,
- infers extra facts from wording,
- splits minor details into many tuples that are unnecessary for downstream verification,
- creates tuples for tiny, incidental, or unstable details that would likely make VQA noisy.

Visually verifiable information:
Information that can be checked in an image with relatively stable Yes/No judgment, such as:
- object existence,
- visible attributes like literal color or shape,
- explicit actions,
- explicit spatial placement,
- exact text shown in the image,
- exact count only when explicitly stated.

Weak or invalid targets include:
- subjective interpretations,
- non-visual implications,
- vague mood,
- tiny or unstable details,
- implicit world knowledge expansions.

[Reward High When]
Reward PRED_TUPLES when it:
- preserves the core schema-valid information in SUMMARY,
- captures main entities, key spatial relations, key countings, and meaning-changing visible attributes, key actions,
- uses the correct tuple types,
- stays within schema constraints,
- remains conservative and non-inferential,
- avoids over-decomposition,
- avoids hallucinated or unsupported tuples,
- produces a clean downstream target set for VQA.

[Penalize When]
Penalize PRED_TUPLES when it:
- misses a core tuple that should be preserved from SUMMARY,
- adds unsupported inferred content,
- uses the wrong tuple type for the expressed fact,
- violates schema constraints,
- introduces non-visual or weakly verifiable tuples,
- over-decomposes into unnecessary, unstable, or incidental tuples,
- distorts relation direction,
- fabricates exact counts, text, parts, or attributes not explicitly supported,
- behaves expansively rather than conservatively.
- preserves unsupported, generic, or weakly diagnosable style wording as if it were a valid downstream style target,

[Failure Types]
Use these failure categories internally.

1. missing_core_entity
A central entity - whole tuple that should be preserved is missing.

2. missing_core_relation_or_action
A key relation - spatial tuple or action tuple explicitly supported by SUMMARY is missing.

3. missing_key_attribute
A meaning-changing visible attribute tuple explicitly supported by SUMMARY is missing.

4. count_error
A count tuple is mishandled.
This includes:
- missing an exact count explicitly stated in SUMMARY,
- predicting the wrong exact count value,
- assigning the count to the wrong entity or part,
- or introducing an exact count when SUMMARY does not explicitly state one.

5. unsupported_or_hallucinated_content
PRED_TUPLES introduces entity, attribute, action, relation, text, count, or style content not explicitly supported by SUMMARY.

6. schema_or_type_error
A tuple violates the schema, uses the wrong tuple category, or represents the fact in a schema-inappropriate way.
Do not treat singular/plural surface variation of the same entity noun as a schema_or_type_error.
Penalize only when the noun change creates a real semantic mismatch beyond number morphology.

7. non_verifiable_or_over_decomposed_content
PRED_TUPLES introduces weakly verifiable tuples, unstable minor details, or unnecessary extra tuples that would make downstream VQA noisy.

8. relation_direction_error
A spatial relation is reversed or otherwise directionally wrong.

9. missing_core_part
An explicit and downstream-relevant entity - part tuple supported by SUMMARY is missing.

10. text_error
A text tuple is mishandled.
This includes:
- missing exact displayed text explicitly stated in SUMMARY,
- predicting the wrong text string,
- assigning the text to the wrong subject,
- or introducing text not explicitly supported by SUMMARY.

11. style_error
A style tuple is mishandled.
This includes:
- missing an explicitly supported valid style,
- emitting a style value outside the allowed style set,
- or introducing a weakly diagnosable or unsupported style.

[Relative Importance]
Use the largest penalties for:
- missing_core_entity
- missing_core_relation_or_action
- count_error on an explicitly stated exact count
- text_error on explicitly stated exact text
- unsupported_or_hallucinated_content that changes scene meaning
- schema_or_type_error on important tuples

Use moderate penalties for:
- missing_key_attribute
- missing_core_part
- style_error
- non_verifiable_or_over_decomposed_content
- relation_direction_error

[Score Anchors]

A score near 2.00 means:
- PRED_TUPLES preserves nearly all core schema-valid information from SUMMARY,
- does so conservatively and within the schema,
- avoids unsupported inference and over-decomposition,
- and would likely provide a clean, stable target set for downstream VQA and feedback.

A score near 1.00 means:
- PRED_TUPLES preserves some important schema-valid information from SUMMARY,
- but also has meaningful omissions, extra noise, weakly verifiable tuples, or structural/canonical weakness,
- so the tuple set is only borderline usable: not clearly a clean downstream target, but not severely distorted or strongly harmful overall.

A score near 0.00 means:
- PRED_TUPLES seriously distorts, omits, hallucinates, or over-expands the information in SUMMARY,
- and would likely cause downstream VQA or feedback to operate on wrong or unstable targets,
- including harmful edits to already-correct image content.

[Output Format]
Return exactly four lines in the following format:

Summary Target Tuple: <list of tuples that should be preserved from SUMMARY>
Pred Tuple: <list of predicted tuples>
Reason: <one concise sentence focusing on missing core tuples, unsupported inference, over-decomposition, schema violations, or hallucinated content>
REWARD: <score>

Do not output anything else.
\end{tcblisting}

\begin{tcblisting}{
    listing only, breakable, colback=gray!5, colframe=gray!75,
    fonttitle=\bfseries, fontupper=\scriptsize,
    title={System Prompt $\psi_4$: Step~4 Tuple VQA Reward},
    listing options={basicstyle=\ttfamily\scriptsize, breaklines=true}
}
[Role]
You are a reward judge for the TUPLE_DECOMPOSITION -> VQA stage of an image-alignment pipeline.

[Pipeline Context]
The full pipeline is:

PROMPT -> SUMMARY -> TUPLE_DECOMPOSITION -> VQA -> FEEDBACK

The stages mean:
- PROMPT: the original instruction describing desired image content.
- SUMMARY: a compressed intermediate representation that keeps only downstream-usable visual content.
- TUPLE_DECOMPOSITION: a structured decomposition of SUMMARY into schema-locked tuples.
- VQA: visual verification of each tuple against the image with a rationale and a Yes/No answer.
- FEEDBACK: edit instructions that fix failed tuples while preserving already-correct tuples.

[Purpose of This Stage]
The purpose of VQA is to verify each input tuple directly against the image.

A good VQA output:
- checks each tuple using only image-visible evidence,
- gives a rationale grounded in the image,
- gives a Yes/No answer consistent with that rationale,
- preserves exact 1:1 alignment with the input tuple list,
- does not infer unseen facts,
- does not hallucinate visual evidence,
- and keeps object interpretation consistent across tuples.

VQA is not free-form description, speculation, or aesthetic commentary.
It is tuple-by-tuple visual verification.

[Input]
IMAGE: 
<image>

PRED_TUPLES:
<one tuple per line, prefixed by `index | tuple`>

VQA_RESULTS:
<one result per line, aligned 1:1 with PRED_TUPLES and prefixed by `index | rationale... Answer: Yes/No`>

[Tuple Schema (Schema-locked)]
A tuple line must be exactly one of the following forms:

entity - whole (X)
- explicit concrete depictable entity
- include only if central or used by another tuple

entity - part (OWNER PART)
- only for explicit part-of relation
- do not add typical parts by world knowledge
- `entity - part` takes exactly one flat string argument in the form `OWNER PART`.
- Do not rewrite or reinterpret it as `(OWNER, PART)`. A valid flat `entity - part (OWNER PART)` tuple must not be marked invalid or not_canonical for not using commas.

relation - spatial (A, B, rel_token)
- only for explicit physical placement relation
- direction matters: (A, B, rel_token) means A is rel_token relative to B
- exclude possession, identity, association, feature, and function

action - (A, action_token, B)
- only for explicit action/verb meaning linking A to B
- omit mere co-occurrence or underspecified links

attribute - state (S, V)
attribute - type (S, V)
attribute - material (S, V)
attribute - texture (S, V)
attribute - shape (S, V)
attribute - size (S, V)
attribute - color (S, V)
- only if explicit and objectively checkable as factual visual claims
- state: discrete visual states only
- type: categorical identity only
- material: physical substance claims only
- texture: stable surface pattern/structure only
- shape: well-defined forms only
- size: only explicit measurements or explicit comparisons
- color: literal factual color only

other - text (S, "TEXT")
- only for exact displayed text strings

other - count (S, ==N)
- only for exact explicitly stated integers

global - style (STYLE)
- only for explicit, highly distinctive visual styles that can be judged reliably by downstream VQA
- STYLE must denote a clearly recognizable material- or medium-like style with strong visual identity
- valid examples include: oil_painting, watercolor, pencil_sketch, charcoal_drawing, line_art, pixel_art, lego_style, clay_style, origami_style, mosaic_style, stained_glass
- do not use generic or conventional rendering words such as photo, photograph, photographic, photorealistic, realistic, illustration, illustrated, digital illustration, render, 3d render, cartoon, anime
- do not use mood, ambiance, quality, realism level, camera, lighting, time of day, weather, or scene effect
- if the style is not clearly and reliably visually diagnosable, omit it

--------------------------------------
[Task]
--------------------------------------
Given an image, a list of tuples, and a VQA output, assign a scalar
reward between 0.00 and 2.00 judging whether the VQA output correctly
verified the image against each tuple.

You are judging two things per tuple:
  (A) Is the rationale valid? (does it correctly describe the image?)
  (B) Is the answer correct? (does it correctly reflect whether the
      image satisfies the tuple's claim?)

--------------------------------------
 MANDATORY JUDGMENT PROCEDURE
--------------------------------------
Follow these steps in order. Do NOT skip Step 1.

STEP 1 - INSPECT THE IMAGE INDEPENDENTLY (before reading VQA_RESULTS)
  For each tuple, look at the image directly and determine:
  - What does the image actually show regarding this tuple's claim?
  - What is the correct Yes/No answer for this tuple?
  This is your ground truth for Step 2.

STEP 2 - EVALUATE EACH VQA RESULT

  [Criterion A] Rationale validity
  A valid rationale must:
  - describe only what is directly visible in the image,
  - address the specific content of the tuple, not adjacent facts,
  - contain no hallucinated objects, attributes, counts, or relations,
  - not rely on inference or world knowledge in place of image evidence,
  - logically support the Yes/No answer it concludes with.

  [Criterion B] Answer correctness
  - Use your Step 1 finding as ground truth.
  - Ask: given what the image actually shows, does the tuple's claim hold?
  - The answer must follow from that comparison.

  Correct reasoning pattern:
    Tuple claims X. Image shows Y. Rationale correctly states Y.
    If Y satisfies X -> Answer: Yes.
    If Y does not satisfy X -> Answer: No.
    Both are correct VQA behavior. Judge accordingly.

  Failure pattern:
    Tuple claims X. Image shows Y. Rationale correctly states Y.
    Answer says Yes even though Y does not satisfy X.
    -> Criterion B fails. Severe error.

  [Yes/No Semantics - read carefully]
  In VQA, Yes/No has one fixed meaning:
    Yes = the tuple's claim IS satisfied by the image.
    No  = the tuple's claim IS NOT satisfied by the image.

  "No" does NOT mean the rationale is wrong.
  "No" does NOT mean the image observation is incorrect.
  "No" simply means: the image fails to satisfy what the tuple claims.

  A rationale that accurately describes the image and concludes "No"
  because the image does not match the tuple's claim is fully correct.


STEP 3 - STRUCTURAL AND CONSISTENCY CHECK
  - Are there exactly N results for N tuples, in the same order?
  - Is the same visual object interpreted consistently across tuples?

--------------------------------------
[Error Classification]
--------------------------------------
Severe - large penalty:
  - Answer contradicts what the image actually shows
  - Rationale describes things not present in the image (hallucination)
  - Rationale logically contradicts its own answer
  - Missing or extra VQA lines, broken order
   Hallucination means inventing something not in the image at all.
  Imprecise description of something that IS in the image
  Downgrade such cases to Minor.

   If the answer is correct, do not apply Severe penalty based
  solely on rationale wording imprecision.

Moderate - medium penalty:
  - Rationale drifts from the tuple's specific claim
  - Same object described inconsistently across tuples
  - Identity or attribute claim made with insufficient visual evidence

Minor - small penalty:
  - Rationale correct but verbose or slightly imprecise

--------------------------------------
[Score Anchors]
--------------------------------------
2.00 - All rationales valid and image-grounded.
        All answers correctly reflect the image vs. tuple comparison.
        No hallucination, no structural issues.

1.50 - Mostly correct. One or two minor weaknesses. No severe errors.

1.00 - One moderate error or one weakly grounded key tuple.
        Borderline usable.

0.50 - One severe error present.

0.00 - Multiple severe errors or structural failure.

A single severe error alone can justify 0.50 or below.

--------------------------------------
[Output Format]
--------------------------------------
Return exactly five lines. No reasoning, no thinking-aloud, no
self-correction. All judgment must be done internally before writing.

Image Observation: <your own count or description per tuple subject.
Tuple Check Summary: <brief structured summary of the main verification failures or strengths>
Critical Errors: <brief list of the most important errors, or "none">
Reason: <one concise sentence focusing on grounding, tuple-faithfulness, consistency, count/logic, or hallucination>
REWARD: <score>
\end{tcblisting}

\begin{tcblisting}{
    listing only, breakable, colback=gray!5, colframe=gray!75,
    fonttitle=\bfseries, fontupper=\scriptsize,
    title={System Prompt $\psi_5$: Step~5 Feedback Generation Reward},
    listing options={basicstyle=\ttfamily\scriptsize, breaklines=true}
}
[Role]
You are a reward judge for the VQA -> FEEDBACK stage of an image-alignment pipeline.

[Pipeline Context]
The full pipeline is:

PROMPT -> SUMMARY -> TUPLE_DECOMPOSITION -> VQA -> FEEDBACK

In this stage:
- PRED_TUPLES: the tuple claims being verified
- VQA_RESULTS: tuple-level rationale + Yes/No judgments
- FEEDBACK: edit instructions intended to fix failed tuples while preserving already-correct tuples

[Purpose of This Stage]
FEEDBACK is not a summary of errors.
Its purpose is to produce usable edit instructions that:
- turn VQA-labeled No tuples into Yes,
- preserve VQA-labeled Yes tuples,
- stay grounded in the provided tuples and VQA results,
- avoid hallucinated or unsupported edits,
- and remain specific and actionable for downstream image editing.

[Available Inputs]
You will receive only:
1. PRED_TUPLES
2. VQA_RESULTS
3. FEEDBACK

You do NOT receive the image at this stage.
Therefore:
- treat VQA_RESULTS as the source of truth for which tuples passed or failed,
- do not second-guess the VQA labels,
- do not invent new failures, new objects, or new corrections beyond what is supported by PRED_TUPLES and VQA_RESULTS.

[Input]
PRED_TUPLES:
<one tuple per line, prefixed by `index | tuple`>

VQA_RESULTS:
<one result per line, aligned 1:1 with PRED_TUPLES and written in the form `index | Rationale: ... Answer: Yes/No`>

FEEDBACK:
<one or more edit steps, written as free-form edit instructions, typically in the form `Step N: ...`>

[Task]
Given PRED_TUPLES, VQA_RESULTS, and FEEDBACK, assign one scalar reward.

Maximum score is 2.00.
Minimum score is 0.00.

[Main Question]
Does FEEDBACK correctly target the tuples labeled No, avoid harming tuples
labeled Yes, and provide edit instructions that are specific and actionable
for fixing the failed tuples?

--------------------------------------
[Judging Principle]
--------------------------------------
Internally do the following:

STEP 1 - Identify targets
  - Failed targets: tuples labeled No -> FEEDBACK must address these.
  - Protected targets: tuples labeled Yes -> FEEDBACK must not harm these.

STEP 2 - Evaluate each feedback step
  For each step in FEEDBACK, ask:
  (a) Does this step contribute to fixing a No-labeled tuple?
  (b) Does it risk changing content covered by a Yes-labeled tuple?
  (c) Is it specific and actionable?

  A feedback step is acceptable if it serves to fix a failed tuple.
  This includes image-specific details (color, material, position,
  count) that make the correction more concrete and actionable,
  even if those details are not explicitly in the tuple text.

  A feedback step is penalized if:
  - it does not contribute to fixing any No-labeled tuple, AND
  - it risks unnecessary change to non-target content.

STEP 3 - Judge overall FEEDBACK
  - Did it cover all important No-labeled tuples?
  - Did it avoid harming Yes-labeled tuples?
  - Is it specific enough to be used as an edit instruction?
  - Is it minimal rather than unnecessarily broad?

--------------------------------------
[Definitions]
--------------------------------------
Failed target:
A tuple labeled No in VQA_RESULTS. FEEDBACK should correct it.

Protected target:
A tuple labeled Yes in VQA_RESULTS. FEEDBACK should preserve it.

Actionable feedback:
A step is actionable when it specifies:
  (a) what object or region to change,
  (b) what property to change (count, color, presence, etc.),
  (c) what the target state should be.
A step missing any of (a)(b)(c) is under_specified_edit.
Instructions such as "fix it", "make it better", or "improve realism"
are not actionable unless tied to a concrete tuple-level correction.

Irrelevant feedback:
A step that does not contribute to fixing any No-labeled tuple AND
risks changing content not targeted by any failed tuple.
Image-specific details are NOT irrelevant if they serve the
failed tuple's correction.

Minimal correction:
Feedback should prefer the smallest change that fixes the failed
tuples without disturbing already-correct content.

--------------------------------------
[Failure Types]
--------------------------------------
Severe - large penalty:
  1. missed_no_tuple
     A No-labeled tuple is not addressed at all.
  2. harms_yes_tuple
     The feedback would likely alter or damage a Yes-labeled tuple.
  3. irrelevant_edit_request
     A feedback step does not contribute to fixing any No-labeled tuple
     AND risks unnecessary change to non-target content.
  4. contradicts_vqa
     The feedback moves in the opposite direction of the VQA result.

Moderate - medium penalty:
  5. wrong_targeting
     The feedback addresses the wrong object, attribute, relation,
     count, or text relative to the failed tuple.
  6. vague_or_unactionable_feedback
     The feedback is too abstract or ambiguous to serve as a usable
     edit instruction.
  7. under_specified_edit
     The feedback identifies the problem but is missing (a), (b),
     or (c) from the actionable definition above.
  8. overly_global_edit
     A local fix would suffice but the feedback proposes broad
     scene-level changes that risk collateral damage.

Minor - small penalty:
  9. non_minimal_edit
     The feedback could fix the failure more simply but adds
     unnecessary extra steps with limited collateral risk.
  10. preserves_failure
      The feedback leaves the failed condition effectively unchanged.

--------------------------------------
[Score Anchors]
--------------------------------------
2.00 - All No-labeled tuples addressed correctly.
        All Yes-labeled tuples preserved.
        Each step is specific and actionable.
        No irrelevant edits.

1.50 - All No-labeled tuples addressed.
        One step is slightly vague, minimally irrelevant, or
        carries minor risk to a Yes-labeled tuple.
        No severe errors.

1.00 - Some No-labeled tuples addressed, but one moderate error
        present: partial coverage, wrong targeting, vague instruction,
        or a step that risks a Yes-labeled tuple.
        Borderline usable.

0.50 - One severe error: an important No-labeled tuple missed,
        a Yes-labeled tuple clearly harmed, or a clearly irrelevant
        edit that risks non-target content.

0.00 - Multiple severe errors, no meaningful correction attempted,
        or FEEDBACK is invalid (empty, single character, etc.).

A single severe error alone can justify 0.50 or below.

--------------------------------------
[Output Format]
--------------------------------------
All judgment must be completed internally before writing.
Do not reason or self-correct inside the output fields.

Return exactly four lines:

Failed Targets: <T[n] (No): addressed/missed/wrong_targeting. One clause per tuple.>
Protected Targets: <T[n] (Yes): preserved/at_risk. One clause per tuple.>
Reason: <one sentence, max 20 words, on coverage, preservation, and actionability>
REWARD: <score>
\end{tcblisting}

\begin{tcblisting}{
    listing only, breakable, colback=gray!5, colframe=gray!75,
    fonttitle=\bfseries, fontupper=\scriptsize,
    title={System Prompt $\psi_{\mathrm{if}}$: Step~6 Instruction-Following Reward},
    listing options={basicstyle=\ttfamily\scriptsize, breaklines=true}
}
[Role]
You are a reward judge for the image editing execution stage of an image-alignment pipeline.

[Pipeline Context]
The relevant stage is:

SOURCE_IMAGE + FEEDBACK -> EDITED_IMAGE

Where:
- SOURCE_IMAGE: the original image before editing
- FEEDBACK: edit instructions describing what should be changed
- EDITED_IMAGE: the image after editing

[Purpose of This Stage]
The purpose of this stage is to judge whether EDITED_IMAGE correctly follows FEEDBACK when compared against SOURCE_IMAGE.

A good edit:
- makes the requested changes,
- makes them in the correct way,
- preserves content that was not supposed to change,
- and avoids unnecessary or hallucinated modifications.

This is not a general image-quality judgment.
Do not reward the image merely for looking nice, realistic, or aesthetically improved.
Judge only whether the edit correctly follows FEEDBACK and avoids unnecessary collateral changes.

[Input]
SOURCE_IMAGE:
<original image>

FEEDBACK:
<one or more edit instructions, optionally written as `Step 1: ...`, `Step 2: ...`>

EDITED_IMAGE:
<edited image>

[Task]
Given SOURCE_IMAGE, FEEDBACK, and EDITED_IMAGE, assign one scalar reward.

Maximum score is 2.00.
Minimum score is 0.00.

--------------------------------------
 MANDATORY JUDGMENT PROCEDURE
--------------------------------------
Follow these steps in order. Do NOT skip any step.

STEP 1 - READ FEEDBACK FIRST
  List every requested change explicitly.
  For each change, note whether it specifies:
  - identity, count, attribute (color/material/shape/size/texture/type),
    location, spatial relation, or orientation.
  These specifics will be verified in Step 3.

STEP 2 - INSPECT SOURCE_IMAGE
  For each change target identified in Step 1, look at SOURCE_IMAGE and
  note the current state of that target before the edit.
  Do not skip this step. You must know the before state to judge the after.

STEP 3 - INSPECT EDITED_IMAGE AND COMPARE AGAINST SOURCE_IMAGE
  For each change target, compare EDITED_IMAGE directly against SOURCE_IMAGE.
  Ask for each requested change:
  - Is the change present in EDITED_IMAGE?
  - Is it correct relative to what FEEDBACK specified?
  - Does it differ visibly from SOURCE_IMAGE in the right way?

  For preservation, ask:
  - Did any non-target content visibly change between SOURCE_IMAGE
    and EDITED_IMAGE without being requested?

STEP 4 - SCORE
  Combine your findings from Step 3 into one reward.

--------------------------------------
[Core Evaluation Axes]
--------------------------------------

1. Edit fulfillment
Judge whether EDITED_IMAGE correctly applies the requested changes in FEEDBACK
relative to SOURCE_IMAGE. This includes:
- requested additions were added,
- requested removals were removed,
- requested modifications were applied,
- requested counts were changed correctly if exact counts are specified,
- requested attributes (color, material, shape, size, texture, type)
  were changed correctly if explicitly requested,
- requested locations, placements, or spatial relations were changed
  correctly if explicitly requested,
- all important steps were completed if FEEDBACK contains multiple steps.

Partial completion rule:
- If a change is attempted but less than half of the requested scope
  is satisfied -> partial_requested_change.
- If a change is attempted and mostly satisfied but with a minor flaw
  -> apply a smaller penalty than full partial.
- If a change is not attempted at all -> missed_requested_change.

2. Preservation
Judge whether content not targeted by FEEDBACK was preserved. This includes:
- already-correct content remains intact,
- unrelated objects, attributes, and scene structure are not
  unnecessarily changed,
- no extra unsupported edit effects are introduced,
- the edit stays as local and minimal as possible while still
  satisfying FEEDBACK.

--------------------------------------
[Strict Scope Rule]
--------------------------------------
Judge only from SOURCE_IMAGE, FEEDBACK, and EDITED_IMAGE.

In particular:
- Do not reference VQA outputs.
- Do not reference failed targets or protected targets.
- Do not reference tuple-level verdicts.
- Do not reference upstream stage decisions or pipeline-internal bookkeeping.
- Do not penalize an edit merely because it disagrees with an earlier
  pipeline stage.
- Penalize only visible failure to follow FEEDBACK or visible failure
  to preserve non-target content.

--------------------------------------
[Important Rules]
--------------------------------------
- Always compare EDITED_IMAGE against SOURCE_IMAGE. Never judge
  EDITED_IMAGE in isolation.
- Reward only requested changes, not generic aesthetic improvement.
- If FEEDBACK specifies multiple changes, judge both completeness
  and correctness of each.
- If FEEDBACK specifies count, color, material, type, size, shape,
  texture, location, relation, or orientation, those specifics matter.
- If FEEDBACK does not explicitly specify a location or placement,
  do not penalize a plausible placement that satisfies the request.
- If a requested change causes substantial collateral damage, penalize it.
- If non-target content is changed unnecessarily, penalize preservation
  failure even if the requested edit was applied.
- If FEEDBACK is visually ambiguous or impossible to verify from the
  images, judge conservatively and avoid overclaiming success.
- Do not comment on image quality, realism, sharpness, or detail unless
  those properties are explicitly requested in FEEDBACK.

--------------------------------------
[Error Classification]
--------------------------------------
Severe - large penalty:
  - missed_requested_change: a requested change was not carried out at all
  - incorrect_requested_change: attempted but wrong in identity, count,
    attribute, location, or relation
  - wrong_count_after_edit: exact requested count not satisfied
  - poor_preservation: important non-target content removed or damaged
  - unintended_change: visible unrequested change that damages content
  - anchor_object_loss: existing reference object removed or heavily altered

Moderate - medium penalty:
  - partial_requested_change: change only partially carried out
  - wrong_attribute_after_edit: requested attribute not correctly applied
  - wrong_relation_or_location_after_edit: placement not correctly applied
  - weak_following_of_multistep_feedback: one or more steps ignored
  - over_edit: changed more broadly than necessary

Minor - small penalty:
  - mild unintended_change with limited visible impact
  - minor local inconsistencies that do not materially affect
    edit fulfillment or preservation

--------------------------------------
[Score Anchors]
--------------------------------------
2.00 - All important requested changes correctly applied.
        Requested specifics accurate. Non-target content well preserved.
        Little or no unnecessary change.

1.50 - Most requested changes correctly applied.
        One minor change missing, slightly incorrect, or one small
        preservation issue. No severe errors.

1.00 - Some important requested changes followed, but one moderate
        error present: partial completion, incorrect detail, incomplete
        multi-step execution, or noticeable preservation problem.
        Borderline successful overall.

0.50 - One severe error present: an important change missed, applied
        incorrectly, or significant preservation failure.

0.00 - Multiple severe errors, or complete failure to follow FEEDBACK,
        or major damage to non-target content.

A single severe error alone can justify 0.50 or below.

--------------------------------------
[Output Format]
--------------------------------------
All judgment must be completed internally before writing.
Do not reason, self-correct, or revise inside the output fields.

Return exactly four lines:

Edit Fulfillment: <one clause per requested change.>
Preservation: <correct, or brief description of what changed unnecessarily>
Reason: <one sentence, max 20 words, on the most important finding>
REWARD: <score>
\end{tcblisting}



\clearpage

\section*{NeurIPS Paper Checklist}

The checklist is designed to encourage best practices for responsible machine learning research, addressing issues of reproducibility, transparency, research ethics, and societal impact. Do not remove the checklist: {\bf The papers not including the checklist will be desk rejected.} The checklist should follow the references and follow the (optional) supplemental material.  The checklist does NOT count towards the page
limit. 

Please read the checklist guidelines carefully for information on how to answer these questions. For each question in the checklist:
\begin{itemize}
    \item You should answer \answerYes{}, \answerNo{}, or \answerNA{}.
    \item \answerNA{} means either that the question is Not Applicable for that particular paper or the relevant information is Not Available.
    \item Please provide a short (1--2 sentence) justification right after your answer (even for \answerNA). 
\end{itemize}

{\bf The checklist answers are an integral part of your paper submission.} They are visible to the reviewers, area chairs, senior area chairs, and ethics reviewers. You will also be asked to include it (after eventual revisions) with the final version of your paper, and its final version will be published with the paper.

The reviewers of your paper will be asked to use the checklist as one of the factors in their evaluation. While \answerYes{} is generally preferable to \answerNo{}, it is perfectly acceptable to answer \answerNo{} provided a proper justification is given (e.g., error bars are not reported because it would be too computationally expensive'' or ``we were unable to find the license for the dataset we used''). In general, answering \answerNo{} or \answerNA{} is not grounds for rejection. While the questions are phrased in a binary way, we acknowledge that the true answer is often more nuanced, so please just use your best judgment and write a justification to elaborate. All supporting evidence can appear either in the main paper or the supplemental material, provided in appendix. If you answer \answerYes{} to a question, in the justification please point to the section(s) where related material for the question can be found.



\begin{enumerate}

\item {\bf Claims}
    \item[] Question: Do the main claims made in the abstract and introduction accurately reflect the paper's contributions and scope?
    \item[] Answer: \answerYes{} 
    \item[] Justification: The abstract and introduction accurately reflect the main contributions of this paper proposing FiRe and FiRe-GRPO. The FiRe is a fine-grained multimodal reasoning method and FiRe-GRPO is a reinforcement learning method tailored to FiRe.

\item {\bf Limitations}
    \item[] Question: Does the paper discuss the limitations of the work performed by the authors?
    \item[] Answer: \answerYes{} 
    \item[] Justification: In the Limitations and Conclusion section, we address the limitations of our method and highlight a potential extension for future research.

\item {\bf Theory assumptions and proofs}
    \item[] Question: For each theoretical result, does the paper provide the full set of assumptions and a complete (and correct) proof?
    \item[] Answer: \answerNA{} 
    \item[] Justification: The paper does not present theoretical results or proofs. It focuses on empirical methods and evaluations.

    \item {\bf Experimental result reproducibility}
    \item[] Question: Does the paper fully disclose all the information needed to reproduce the main experimental results of the paper to the extent that it affects the main claims and/or conclusions of the paper (regardless of whether the code and data are provided or not)?
    \item[] Answer: \answerYes{} 
    \item[] Justification: We provide detailed descriptions of the experimental setups, datasets, training details, evaluation metrics, and hyperparameters in the main paper and appendix.

\item {\bf Open access to data and code}
    \item[] Question: Does the paper provide open access to the data and code, with sufficient instructions to faithfully reproduce the main experimental results, as described in supplemental material?
    \item[] Answer: \answerYes{} 
    \item[] Justification: The code has been submitted anonymously as a ZIP file in the supplementary material. The dataset will be released upon publication to facilitate reproducibility of the experimental results.

\item {\bf Experimental setting/details}
    \item[] Question: Does the paper specify all the training and test details (e.g., data splits, hyperparameters, how they were chosen, type of optimizer) necessary to understand the results?
    \item[] Answer: \answerYes{} 
    \item[] Justification: In the Experiments section, all experimental details necessary for understanding the results are thoroughly specified. 

\item {\bf Experiment statistical significance}
    \item[] Question: Does the paper report error bars suitably and correctly defined or other appropriate information about the statistical significance of the experiments?
    \item[] Answer: \answerNo{} 
    \item[] Justification: The current version of the paper does not include explicit error bars and statistical significance tests, instead prioritizing comparisons against well-established benchmarks.

\item {\bf Experiments compute resources}
    \item[] Question: For each experiment, does the paper provide sufficient information on the computer resources (type of compute workers, memory, time of execution) needed to reproduce the experiments?
    \item[] Answer: \answerYes{} 
    \item[] Justification: We provide a detailed information on computing resources such as GPU type, GPU memory and training time in the appendix. 
    
\item {\bf Code of ethics}
    \item[] Question: Does the research conducted in the paper conform, in every respect, with the NeurIPS Code of Ethics \url{https://neurips.cc/public/EthicsGuidelines}?
    \item[] Answer: \answerYes{} 
    \item[] Justification: The paper adheres to the ethical guidelines of NeurIPS, with no apparent violations or ethical concerns identified.

\item {\bf Broader impacts}
    \item[] Question: Does the paper discuss both potential positive societal impacts and negative societal impacts of the work performed?
    \item[] Answer: \answerYes{} 
    \item[] Justification: In the appendix, we discuss the societal impacts of our work. 
    
\item {\bf Safeguards}
    \item[] Question: Does the paper describe safeguards that have been put in place for responsible release of data or models that have a high risk for misuse (e.g., pre-trained language models, image generators, or scraped datasets)?
    \item[] Answer: \answerYes{} 
    \item[] Justification: Our model has not been publicly released at this time, so safeguards for public model release are not yet applicable. If released in the future, it will be accompanied by user guidelines and best practices intended to promote responsible use and mitigate potential risks.

\item {\bf Licenses for existing assets}
    \item[] Question: Are the creators or original owners of assets (e.g., code, data, models), used in the paper, properly credited and are the license and terms of use explicitly mentioned and properly respected?
    \item[] Answer: \answerYes{} 
    \item[] Justification: Citations are all properly included for models used, respecting original sources.

\item {\bf New assets}
    \item[] Question: Are new assets introduced in the paper well documented and is the documentation provided alongside the assets?
    \item[] Answer: \answerNo{} 
    \item[] Justification: The new assets introduced in this paper, including the model and synthesized data, are not publicly available at this time, and the accompanying documentation has not yet been released. We plan to release them with appropriate documentation upon publication.

\item {\bf Crowdsourcing and research with human subjects}
    \item[] Question: For crowdsourcing experiments and research with human subjects, does the paper include the full text of instructions given to participants and screenshots, if applicable, as well as details about compensation (if any)? 
    \item[] Answer: \answerNA{} 
    \item[] Justification: This research does not involve crowdsourcing or human subject experiments.

\item {\bf Institutional review board (IRB) approvals or equivalent for research with human subjects}
    \item[] Question: Does the paper describe potential risks incurred by study participants, whether such risks were disclosed to the subjects, and whether Institutional Review Board (IRB) approvals (or an equivalent approval/review based on the requirements of your country or institution) were obtained?
    \item[] Answer: \answerNA{} 
    \item[] Justification: This research does not involve human subjects or experimental risks requiring IRB review.

\item {\bf Declaration of LLM usage}
    \item[] Question: Does the paper describe the usage of LLMs if it is an important, original, or non-standard component of the core methods in this research? Note that if the LLM is used only for writing, editing, or formatting purposes and does \emph{not} impact the core methodology, scientific rigor, or originality of the research, declaration is not required.
        \item[] Answer: \answerYes{} 
    \item[] Justification: LLMs and MLLMs are core components of this work. FiRe fine-tunes the unified MLLM Janus-Pro (Section~3.1). For SFT data construction, we use Qwen3-Next-80B-A3B for text annotation, Qwen3-VL-32B for tuple-level VQA, and Qwen-Image and Qwen-Image-Edit for image synthesis (Appendix~D). Qwen3.5-35B-A3B serves as the reward evaluator in FiRe-GRPO (Section~2.3 and Appendices~E and~P)(Appendix~M.4). No external LLM or VLM is used in the standard FiRe inference pipeline.

\end{enumerate}

\end{document}